\documentclass[times,1p]{elsarticle}

\usepackage{comment}
\usepackage{graphicx}
\usepackage{amsmath}
\usepackage{amssymb}
\usepackage{subfigure}
\usepackage{eurosym}
\setcitestyle{authoryear,open={(},close={)}}
\usepackage{mathabx}
\usepackage{siunitx}
\usepackage[table,xcdraw]{xcolor}
\definecolor{green}{rgb}{0,0.8,0}
\usepackage{algorithm,algpseudocode}
\usepackage{caption}
\usepackage{url}
\usepackage{lineno}
\usepackage{framed}
\usepackage{numprint}

\usepackage[symbol]{footmisc}

\newcommand{\infint}{\int_{-\infty}^{\infty}}
\newcommand{\infmint}{\int_{\mathbb{R}^q}}
\newcommand{\dset}{\mathcal{D}}

\newcommand{\datax}{\textbf{x}}
\newcommand{\datap}{\textbf{d}}
\newcommand{\w}{\boldsymbol{\theta}}
\newcommand{\g}{\textbf{g}}
\newcommand{\pyxd}{p(y|\datax, \dset)}
\newcommand{\ymean}{\hat{y}_{\text{MP}}(\datax)}
\newcommand{\BM}{BMVI }
\newcommand{\wmp}{\boldsymbol{\theta}_{\text{MP}}}
\newcommand{\xnew}{\textbf{x}_{\text{new}}}
\newcommand{\yvar}{\sigma_{\dset}^2(\datax)}
\DeclareMathOperator*{\argmax}{argmax}
\DeclareMathOperator*{\argmin}{argmin}
\newcommand{\fracset}{\textbf{f}}

\usepackage{setspace}

\providecommand{\keywords}[1]
{
  \small	
  \textbf{\textit{Keywords---}} #1
}

\title{Utilizing remote sensing data in forest inventory\\ sampling via Bayesian optimization}
\author{Jonne Pohjankukka$^{a,}$\footnote[1]{Corresponding author.}\footnote[0]{E-mail address: jjepoh@utu.fi (J. Pohjankukka)}, Sakari Tuominen$^b$ and Jukka Heikkonen$^a$ \\[6pt] $^a$Department of Future Technologies, University of Turku\\ Vesilinnantie 5, FI-20500 Turku\\$^b$Natural Resources Institute Finland (LUKE)\\ Latokartanonkaari 9, FI-00790 Helsinki}
\date{}

\makeatletter
\def\ps@pprintTitle{%
   \let\@oddhead\@empty
   \let\@evenhead\@empty
   \let\@oddfoot\@empty
   \let\@evenfoot\@oddfoot
}
\makeatother

\begin{document}
\begin{abstract}
In large-area forest inventories a trade-off between the amount of data to be sampled and the costs of collecting the data is necessary. It is not always possible to have a very large data sample when dealing with sampling-based inventories. It is therefore necessary to optimize the sampling design in order to achieve optimal population parameter estimation. On the contrary, the availability of remote sensing (RS) data correlated with the forest inventory variables is usually much higher. The combination of RS and the sampled field measurement data is often used for improving the forest inventory parameter estimation. In addition, it is also reasonable to study the utilization of RS data in inventory sampling, which can further improve the estimation of forest variables. In this study, we propose a data sampling method based on Bayesian optimization which uses RS data in forest inventory sample selection. The presented method applies the learned functional relationship between the RS and inventory data in new sampling decisions. We evaluate our method by conducting simulated sampling experiments with both synthetic data and measured data from the Aland region in Finland. The proposed method is benchmarked against two baseline methods: simple random sampling and the local pivotal method. The results of the simulated experiments show the best results in terms of MSE values for the proposed method when the functional relationship between RS and inventory data is correctly learned from the available training data.
\end{abstract}

\maketitle

\keywords{forest inventory, remote sensing, spatial data sampling, Bayesian optimization, geographic information science, machine learning}
\doublespacing
\section{Introduction}
Large area forest inventories at regional and national level are typically based on sampled field observations measured from sample plots. The sampling intensity is dependent on the size of the inventory area, desired accuracy of the inventory data and the resources available for measuring the data. The sampled data should be representative enough to cover the variation of the significant variables, such as the volume of growing stock and main tree species in the inventory area, in order to allow the estimation of these variables at national and regional level. Increasing the number of field observations generally improves the accuracy of the inventory data, but on the other hand, the measurement of field data is the most significant cost factor in sampling-based forest inventories. Thus, the selected inventory design is always a trade-off between the desired accuracy of inventory data and the available resources.

A wide range of research has been conducted in the field of statistical sampling theory \citep[see e.g.][]{Fuller2009,Kangas2006,Cochran77,Loetsch1964}, which are now applied in many large-scale inventories. For example, simple random sampling, Poisson sampling, stratified sampling, systematic sampling, two-stage sampling and cluster sampling are applied in forest inventories. For details on the sampling methods see e.g. \citep[][]{Fuller2009,Kondo2014}. Methods such as systematic or clustered sampling are common in forest inventories, because the weights of individual sample plots are constant, which makes their application straightforward in monitoring the forest resources over consecutive inventories.  Typical characteristic among the traditional sampling methods, such as random, systematic or clustered sampling, is that they focus mainly on determining the sampling strategy using the information available in the response variable itself. However, the efficiency of sampling design can be improved by using auxiliary data such as remote sensing (RS) data, which as such is not accurate enough for the inventory task but which can be used for enhancing the sampling efficiency by e.g. weighting the areas represented by each sample plot. The main prerequisite for the use of auxiliary data is that there is sufficient correlation between the auxiliary data and actual variables of interest, which typically is the case between RS data and forest inventory variables \citep[e.g.][]{Wallner2017,McRoberts2007,PULITI2017115,Abegg2017,Kangas2018,Joanne2016,Saukkola2019}.  

As a starting point we can assume that we have a field sample where we have a finite number of data observations available for example describing the volume of growing stock at corresponding geographical locations, which we use for estimating the total volume of the inventory area. Furthermore, we can assume that we have additional RS (e.g. satellite) data available describing the surface reflectance throughout the entire research area. In order to improve the accuracy of the inventory data we want to sample data points from new geographical locations. It is then possible to improve the sampling by building a prediction model between the response variable (e.g. volume) and the RS data, and then utilize this relationship in new sampling decisions. That is, we take advantage of the combined information provided by both the response variable itself and the auxiliary RS data.

Recent examples in Swedish forest inventory utilizing auxiliary information in sampling decisions have been presented e.g. in the studies by \citep{Grafstrom2012,Grafstrom2013,Grafstrom2014,Grafstrom2017}. Also, in the works of \citep{Raty2018} and \citep{Raty2019} the authors have applied the local pivotal method in national forest inventory (NFI) using Southern Finland as the test area. The local pivotal method produces sampling decisions in a stochastic manner while trying to avoid similar data points to be included into the data set, in order to produce a spatially well-balanced data set. The results showed significant improvement in estimation accuracy with the utilization of local pivotal method to the NFI data, showing that the auxiliary data can indeed improve the sampling decisions.

In this article, we propose a sampling method for estimating forest inventory variable population parameters based on the well-founded literature on Bayesian optimization \citep[see e.g.][]{Lawrence2009,Nguyen2017,Wang2017,Xu2011}, which also utilizes auxiliary data in sampling decisions. Our proposed method takes advantage of the auxiliary data by building a probabilistic prediction model and basing the sampling decisions on the posterior predictive distribution of the response variable. Our objective is to provide an alternative method to be used in environmental data sampling, in order to improve the representativeness of the data and to minimize the required sample size. In other words, via the proposed method we aim to improve the field data sampling decisions by selecting sample data points containing maximum information content on the underlying phenomena in the data. Note that although we propose our method for improving sampling in forest inventories, the method itself is not limited to this context and can be applied in a general case. As the experimental results show in later sections, if we can produce a well-placed prediction model for the auxiliary and the forest inventory data, then our proposed method can improve the sampling decisions in this context.  

The motivation for our proposed method emerges from the application of RS and field sample data in the forest inventories. Management of forest resources requires predictions e.g. on the distribution of tree species, state of forests, soil conditions for trafficability assessment etc. \citep[see e.g.][]{Pennanen2003,Pohjankukka2014a,Pohjankukka2014b,Pohjankukka2015} in the form of thematic maps. Information gain from these data-based approaches will be utilized both in strategic and operative plannings in forestry. Since the thematic maps are based on a finite number of field samples it is of great of importance to select these samples as optimally as possible. Hence, sampling methods offering possible improvements to the inventory accuracy are welcomed.

In general, NFIs typically cover hundreds of variables, of which information is recorded on NFI sample plots. These variables typically cover, among others, the trees (living or dead), site type, forest health as well as
variables related to biodiversity or ecological value. However, it is not feasible to optimize the sampling design for all variables of interest. For example in Finnish NFI, the current systematic cluster sampling design is optimized for producing unbiased estimates of the total volume of growing stock as well as the volumes of main tree species at regional and national level, and the same sampling design will be used also for all other variables recorded in NFI. Furthermore, plot sampling typically is not best suited for acquiring information of rare phenomena such as rare tree species or plant communities. Other sampling systems such as line or strip sampling would serve better that purpose.

In what follows, section \ref{Section::Data_and_research_area} describes the data sets and research area, section \ref{Section::Methods} presents the technical part of the manuscript, and finally sections \ref{Section:analysis_and_results}, \ref{Section::Discussion} and \ref{Section::conclusions} present the results, discussion and conclusions respectively. To summarize, in this work we aim to answer the following research question:

\begin{framed}
\begin{itemize}
\item Given a learned prediction model between remote sensing and known forest inventory data, can we improve future forest inventory population parameter estimations by utilizing the model's prediction uncertainty in new sampling decisions?  
\end{itemize}
\end{framed}

\section{Materials}\label{Section::Data_and_research_area}
We conducted empirical analyses with the proposed sampling method using both real world forest inventory and RS data in combination with synthetically generated Gaussian mixture model data.

\begin{figure}[b!]
\centering
\resizebox{\textwidth}{!}{
\includegraphics{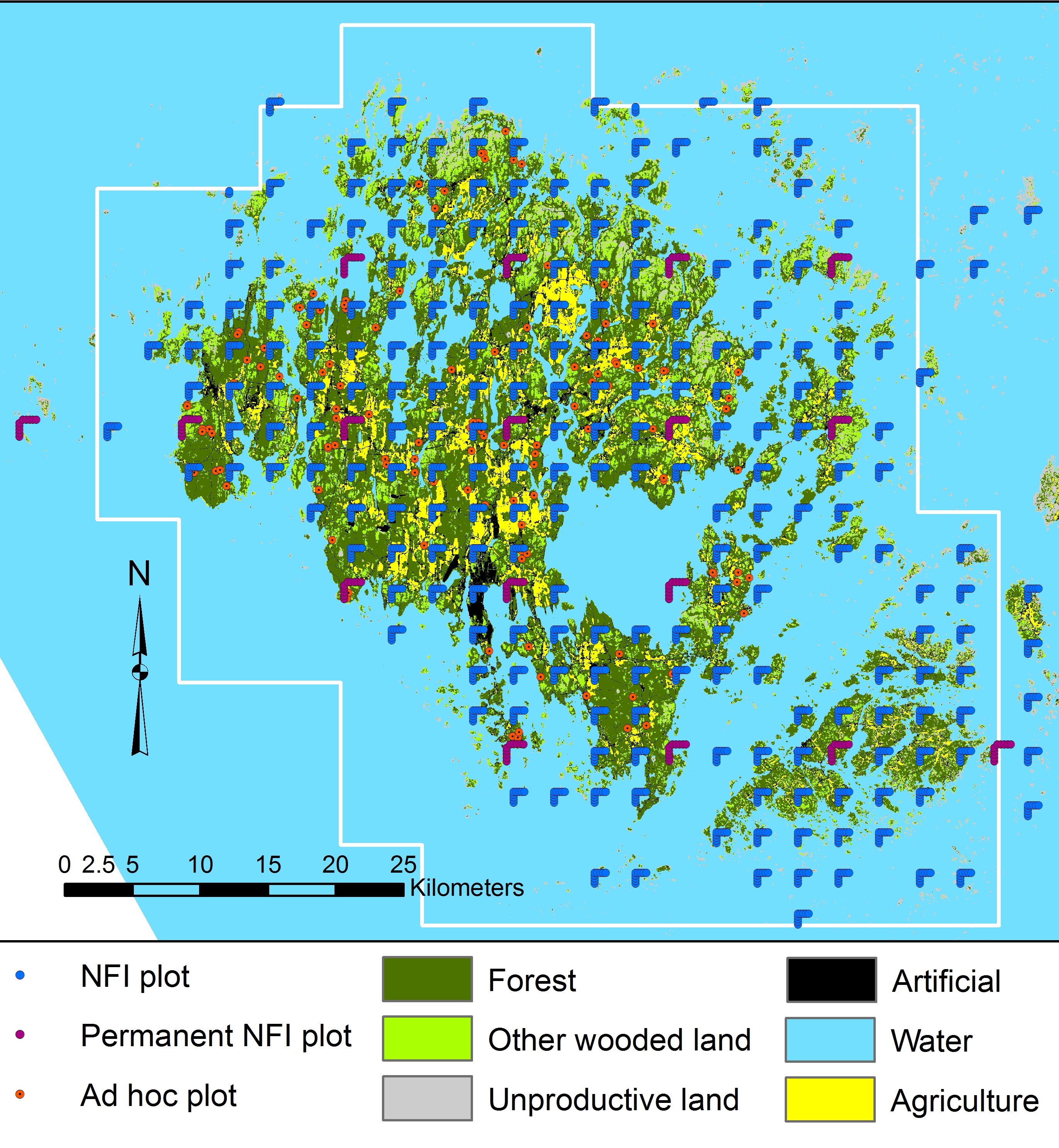}
}
\caption{Map of the sampling layout in Aland. Background land cover map is based on NFI and topographic data provided by LUKE. The ALS coverage area is marked with white borderline.}
\label{Figure::Aland_map}
\end{figure}

\subsection{Study area and field data}
The real world research data was collected from the archipelago province of Aland (lat. \ang{60}11'58.38" N, long. \ang{20}22'16.22" E) in Finland. The data set consisted from a set of airborne laser scanning (ALS), aerial imagery and reference data measured in the field. The ALS and aerial imagery data were used as predictor/input data to generalize the field reference data, i.e. response data, over a larger area. The total area covered by both ALS and aerial imagery data was approximately 346000 ha, but a large part of it was sea area. The field data was mainly composed of 11th Finnish national forest inventory (NFI11) sample plots allocated on the basis of systematic cluster sampling. In the study area sample plot clusters were established in a grid of 3 km by 3 km, and each cluster consisted of 9 sample plots in L-shaped form, having 200 meters distance between plots (see Figure \ref{Figure::Aland_map}). In addition to these sample plot clusters, permanent clusters established in 9th NFI \citep[NFI9,][]{Tomppo2011} were remeasured. A total of 349 NFI plots were measured in forestry land based on systematic sampling. 

In order to cover all forest types present in the inventory area, the inventory area was stratified into 196 strata based on ALS and aerial image features \citep{Pohjankukka2018}. The representativeness of the systematic sample was examined in relation to these strata, and additional field plots were allocated to those strata that were underrepresented or missing in the systematic sample. The additional (ad hoc) sample plots were selected as a random sample from each underrepresented or missing stratum, and the additional plots were not clustered. Altogether 126 plots were allocated to the strata underrepresented or missing among the systematic sample, bringing the total number of field plots to 475 in forestry land. A detailed description of the allocation and measurement of the additional field sample plots is presented in \citep{Pohjankukka2018}.

The sample plots were measured as restricted relascope plots with a basal area factor 1 and maximum radius 9 meters. For each sample plot, tree and stand level variables were recorded in accordance with NFI field guide and nomenclature \citep{Metsantutkimus2009}. The field variables that were applied for testing different sampling strategies in this study were volume of total growing stock and volumes per following tree species groups: pine, spruce and broadleaf trees. In practical forest inventories the amount of growing stock and proportions of tree species are typically the most important stand variables, especially for forest management \citep[e.g.][]{Haara2019}.

\subsection{Remote sensing data}\label{Section::RS_data}

The ALS and aerial imagery data contained a total of 154 variables covering point cloud features from ALS data as well as spectral and textural features from aerial imagery. The following features were extracted from ALS point cloud data from an area representing each 9 meter radius sample plots \citep{Naesset2004,Packalen2006,Packalen2008}:

\begin{enumerate}
    \item Average, standard deviation and coefficient of variation of height above ground ($H$) for canopy returns, separately from first (f) and last (l) returns (havg[f/l], hstd[f/l], hcv[f/l]). 
    \item $H$ at which $p$\% of cumulative sum of $H$ of canopy returns is achieved ($H_p$) (h$p$[f/l], $p$ is one of 0, 5, 10, 20, 30, 40, 50, 60, 70, 80, 85, 90, 95 and 100).
    \item Percentage of canopy returns having $H\geq$ than corresponding $H_p$ ($p$[f/l], $p$ is one of 20, 40, 60, 80, 95).
    \item Canopy densities corresponding to the proportions of points above fraction no. 0, 1, ..., 9 to a total number of points (d0, d1, ..., d9).
    \item (a) Ratio of first canopy returns to all first returns (vegf), and (b) Ratio of last canopy returns to all last returns (vegl). 
    \item Ratio of intensity percentile $p$ to the median of intensity for canopy returns (i$p$[f/l], $p$ is one of 20, 40, 60 and 80).
\end{enumerate}

\noindent The following features were extracted from the aerial image bands from an area representing the size of sample plots:

\begin{enumerate}
    \item Average, standard deviation (std) and coefficient of variation (cv) from each of the four image bands: near-infrared (nir), red (r), green (g), blue (b)
    \item The following multiband transformations \citep[Normalized difference vegetation index, NDVI. See e.g.][]{Yengoh:2015}: 
    \begin{itemize}
        \item NDVI as (nir - r)/(nir + r) 
        \item modified NDVI as (nir − g)/(nir + g) 
        \item nir/r 
        \item nir/g 
    \end{itemize}
    \item Haralick textural features \citep{Haralick73} based on co-occurrence matrices of image band values:
    \begin{itemize}
        \item angular second moment (ASM)
        \item contrast (Contr)
        \item correlation (Corr)
        \item variance (Var)
        \item inverse difference moment (IDM)
        \item sum average (SA)
        \item sum variance (SV)
        \item sum entropy (SE)
        \item entropy (Entr)
        \item difference variance (DV) 
        \item difference entropy (DE)
    \end{itemize}{}
\end{enumerate}

\begin{table*}[t!]
	\footnotesize
	\begin{center}
    \caption{List of the top ten predictor features found for the target variables: growing stock tree volume (all, pine, spruce, broadleaf trees). The Feature ID is a direct reference to a detailed table in the appendix of study \citep{Pohjankukka2018}. In the table $H$ stands for height above ground.}
	\label{table:case_data_sets}	
    	\resizebox{.81\textwidth}{!}{%
		\begin{tabular}{| c | c |}
 			\hline
 			\rowcolor{gray!10}\textbf{Feature ID} & \textbf{Feature description}  \\\hline
 			\rowcolor{gray!25} & \textbf{Volume all trees}  \\\hline
 			134 & texture feature, sum average, ALS based canopy height \\\hline
 			47 & percentage of last canopy returns above 20\% height limit \\\hline
 			45 & percentage of first canopy returns above 90\% height limit \\\hline
 			137 & texture feature, entropy, ALS based canopy height \\\hline
 			129 & texture feature, angular second moment, ALS based canopy height \\\hline
 			133 & texture feature, inverse difference moment, ALS based canopy height \\\hline
 			32 & $H$ at which 100\% of cumulative sum of last canopy returns is achieved ($H_p, p\%$) \\\hline
 			152 & gndvi, transformation from band averages within the pixel windows: nir−g/nir+g \\\hline
 			60 & percentage of last canopy returns having $H \geq$ than corresponding $H_{20}$ \\\hline
 			33 & coefficient of determination of first returned canopy returns \\\hline
 			
 			\rowcolor{gray!25} & \textbf{Volume pine trees}  \\\hline
 			52 & percentage of last canopy returns above 70\% height limit \\\hline
 			140 & texture feature, angular second moment, ALS based intensity \\\hline
 			119 & texture feature, contrast, near-infrared band of CIR imagery \\\hline
 			153 & transformation from band averages within the pixel windows: nir/r \\\hline
 			134 & texture feature, sum average, ALS based canopy height \\\hline
 			145 & texture feature, sum average, ALS based intensity \\\hline
 			94 & texture feature, difference variance, blue band of RGB imagery \\\hline
 			128 & texture feature, difference entropy, near-infrared band of CIR imagery \\\hline
 			36 & ratio of last canopy returns to all last returns \\\hline
 			35 & ratio of first canopy returns to all first returns \\\hline
 			
 			\rowcolor{gray!25} & \textbf{Volume spruce trees}  \\\hline
 			69 & ratio of intensity percentile $20$ to the median of intensity for last canopy returns \\\hline
 			48 & percentage of last canopy returns above 30\% height limit \\\hline
 			71 & ratio of intensity percentile $60$ to the median of intensity for last canopy returns \\\hline
 			68 & ratio of intensity percentile $80$ to the median of intensity for first canopy returns \\\hline
 			94 & texture feature, difference variance, blue band of RGB imagery \\\hline
 			84 & texture feature, coefficient of determination, near-infrared band of CIR imagery \\\hline
 			39 & percentage of first canopy returns above 30\% height limit \\\hline
 			34 & coefficient of determination of last returned canopy returns \\\hline
 			81 & texture feature, coefficient of determination, red band of CIR imagery \\\hline
 			86 & texture feature, contrast, blue band of RGB imagery \\\hline
 			
 			\rowcolor{gray!25} & \textbf{Volume broadleaf trees}  \\\hline
 			145 & texture feature, sum average, ALS based intensity \\\hline
  			66 & ratio of intensity percentile $40$ to the median of intensity for first canopy returns \\\hline
 			65 & ratio of intensity percentile $20$ to the median of intensity for first canopy returns \\\hline
 			70 & ratio of intensity percentile $40$ to the median of intensity for last canopy returns \\\hline
 			148 & texture feature, entropy, ALS based intensity \\\hline
 			143 & texture feature, variance, ALS based intensity \\\hline
 			144 & texture feature, inverse difference moment, ALS based intensity \\\hline
 			59 & percentage of first canopy returns having $H \geq$ than corresponding $H_{95}$ \\\hline
 			38 & percentage of first canopy returns above 20\% height limit \\\hline
 			20 & $H$ at which 5\% of cumulative sum of last canopy returns is achieved ($H_p, p\%$) \\\hline
 			
		\end{tabular}
        }
	\end{center} 
\end{table*}

\begin{figure}[t!]
\centering
\resizebox{.7\textwidth}{!}{
\includegraphics{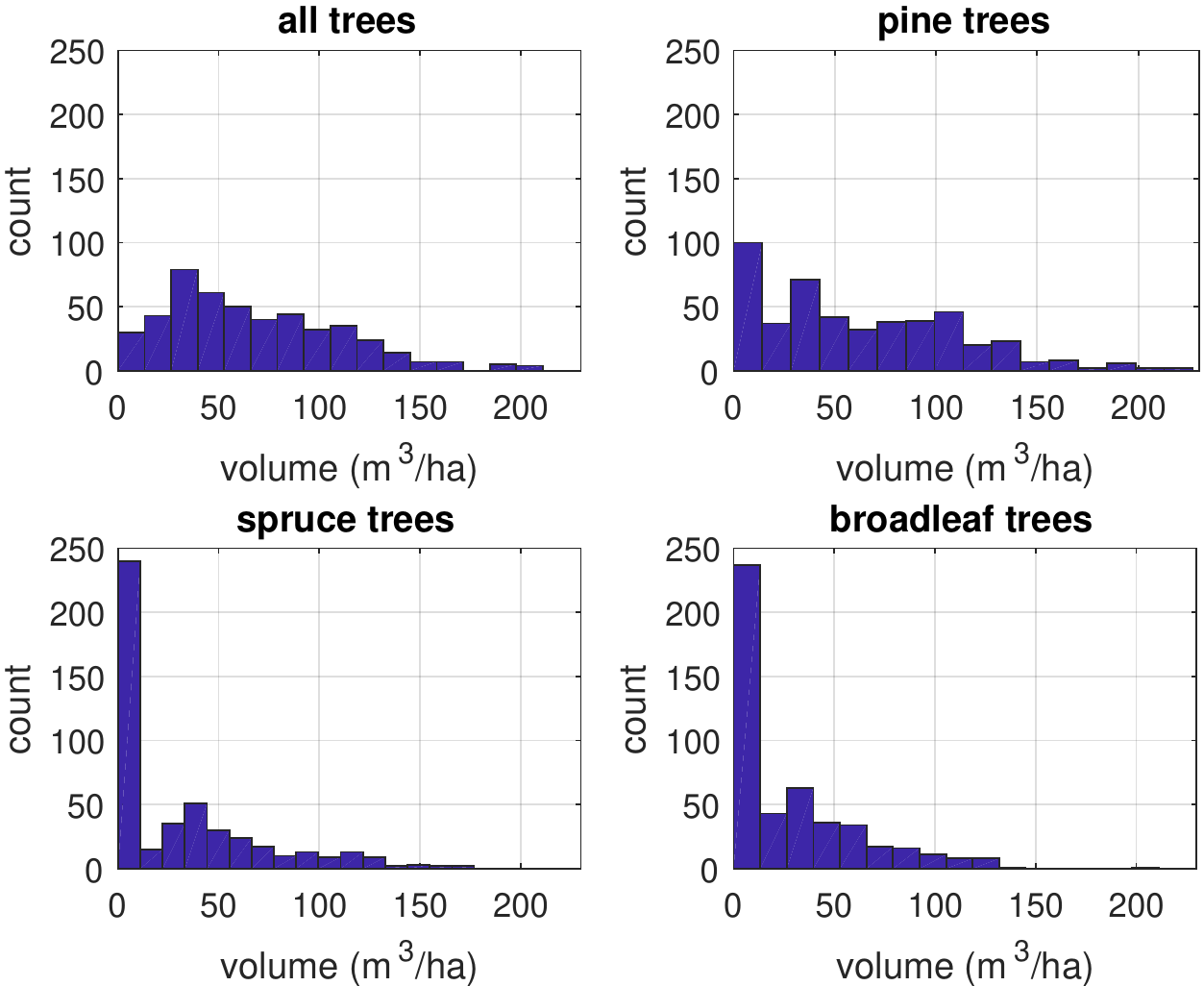}}
\caption{Histograms describing tree volume distributions in the Aland research area. Volumes of 0-10 m$^3$/ha of spruce and broadleaf trees are of high frequency in sample sites, whereas the most frequent volume of all trees in sample sites is 20-30 m$^3$/ha.}
\label{Figure::Aland_volume_distributions}
\end{figure}

\noindent Additionally, height and intensity values of LiDAR points were interpolated into raster format data with similar resolution as aerial imagery for extracting the same textural features as from aerial imagery. A detailed description of the acquisition of RS data is presented in \citep{Pohjankukka2018}.

In this research, we use the ten best features discovered from the ALS and aerial imagery data in the study \citep{Pohjankukka2018} as the auxiliary data for the response variables: volume of growing stock (all trees, pine, spruce and broadleaf) in the field reference data. In the referenced study, approximately ten predictor features were found to be sufficient to achieve optimal prediction performance for the corresponding response variables. We have listed the response variables and their corresponding top ten used predictor features in Table \ref{table:case_data_sets}. Histograms describing the value distributions of the response variables is presented in Figure \ref{Figure::Aland_volume_distributions}. The total number of available data points was 475.


\subsection{Synthetic Gaussian mixture model data}\label{Section::Synthetic_data}
\begin{figure}[t!]
\centering
\resizebox{.6\textwidth}{!}{
\includegraphics{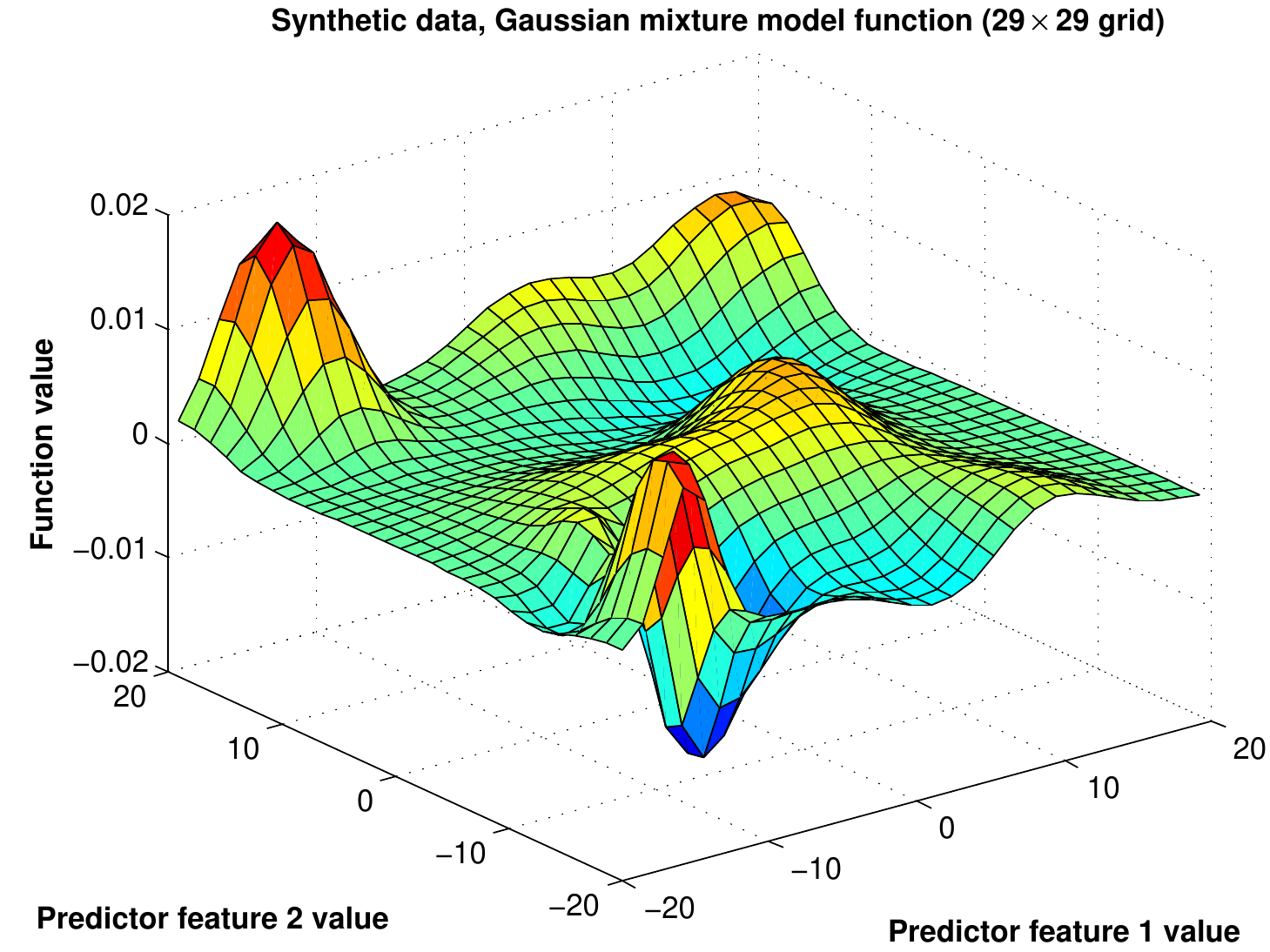}
}
\caption{The synthetically generated random two-dimensional GMM data set. A total of 841 data points were generated consisting from the 29 x 29 grid.}
\label{Figure::synthetic_data}
\end{figure}

The synthetic data set was generated via Gaussian mixture models \citep[GMM, see e.g.][]{Bishop::Pattern_recog_ML_2006} with a two dimensional input feature space. The GMM function $f(\textbf{x})$ was randomly generated with 10 Gaussian clusters. Explicitly put, the relationship between the auxiliary variables $\textbf{x}=(x_1, x_2)$ and response variable $y$ was:
\begin{equation}
    y=f(\textbf{x})=\sum_{i=1}^{10} \mathcal{N}\left(\textbf{x}\,|\,\mu_i, \Sigma_i\right),
\end{equation}
where $\mu_i$ and $\Sigma_i$ are the mean vector and covariance matrix of the corresponding $i$th Gaussian component. A total of 841 data points were generated from this function with the input vector values $x_1, x_2 \in [-20, 20]$ for all $\datax$. We have illustrated the randomly generated data set and the corresponding GMM function in Figure \ref{Figure::synthetic_data}. Note in the figure that the edges (with high magnitude $\datax$) of the plot contain larger variations in the function value. This fact becomes useful for our proposed method in the case of a linear prediction model as we will see in later sections. 

\section{Methods}\label{Section::Methods}
The following notation will be used throughout this and the following sections. A single observation of input or auxiliary predictor variables is denoted as a vector $\datax \in \mathbb{R}^m$ with $m$ distinct features. A corresponding response variable is denoted as $y\in\mathbb{R}$. The pair $\textbf{d}=(\datax, y)$ is treated as a single data point. For example, $\datax$ might contain RS data (e.g. raster pixel information or derived features) on some geographic location and $y$ could contain the average volume of trees in that corresponding location. In this study, $\datax$ and $y$ refer to the auxiliary and response data (forest inventory) as presented in section \ref{Section::Data_and_research_area}. An observed data set is denoted as $\dset=(\mathcal{X}, \mathcal{Y})$, where $\mathcal{X} = \{\datax_1, \datax_2, ..., \datax_N\}$ is the set of $N$ input vectors and $\mathcal{Y}=\{y_1, y_2, ..., y_N\}$ is the set of $N$ realizations of the response variable. A prediction model is denoted as $f(\datax ; \w)$, where $\w \in \mathbb{R}^q$ is a vector of model parameters.


\subsection{Simple random and local pivotal method sampling}
We will compare our proposed sampling algorithm with two sampling methods: simple random sampling \citep[SRS, see e.g.][]{Fuller2009} and local pivotal method sampling \citep[LPM,][]{Deville1998,Grafstrom2012,Saad2016}. To introduce these two methods, we use the symbol $\mathcal{U}$ to denote the population of all possible data points (i.e. all the data that can be sampled) and $\pi_i$ ($0\leq \pi_i \leq 1$) to denote the inclusion probability of data point $\datap_i$. That is, $\pi_i$ is the probability that the $i^{\text{th}}$ data point of population $\mathcal{U}$ will be included into the observed (sampled) data set $\dset \subset \mathcal{U}$. In other words, it is its probability of becoming part of the sample during the drawing of a single sample. In SRS, the inclusion probabilities for all data points are equal, i.e. $\pi_i = \pi_j\;\forall\, i,j$. SRS is a suitable method for many cases and produces an unbiased data set $\dset$, but it can be expensive to implement in field sampling cases like e.g. with geographical data. 

The LPM is a sampling method based on the idea of avoiding the selection of data points that are similar in the feature space $\mathcal{X}$. The point is to select a spatially balanced data $\dset$ from the population $\mathcal{U}$. LPM attempts to select the spatially balanced samples by locally aggregating the inclusion probabilities of neighboring data points, decreasing the likelihood that similar data samples are selected. This for example, is especially useful when we want to acquire a representative sample of geographical data. The LPM starts with an initial inclusion probability set $\Pi = \{\pi_1, \pi_2, ..., \pi_{\lvert\mathcal{U}\rvert}\}$ and proceeds by iteratively updating pairs of inclusion probabilities $(\pi_i, \pi_j)$, so that the sampling outcome is decided for at least one of the two corresponding data points in each iteration.

\begin{algorithm}[b!]
  \caption{Pseudocode for LPM}
  \begin{algorithmic}[1]
    \Require $\mathcal{U}, \Pi$ \Comment{The population data and set of initial inclusion probabilities}
    \Ensure $\dset$ \Comment{The returned sample data}
    \State set $\dset=\emptyset$ and $\mathcal{U}^* = \mathcal{U}$
    \While {$\lvert \mathcal{U}\rvert > 0$} \Comment{Repeat until sampling decision is made for all the data}
        \State Randomly select a data point $\datap_i$ from set $\mathcal{U}$ with uniform probability
        \State Set $\datap_j = \argmin_{\datap \in \mathcal{U}\setminus \datap_i} e(\textbf{x}, \textbf{x}_i)$ \Comment{find the nearest neighbor}
        \State Update the inclusion probabilities $\pi_i, \pi_j\in \Pi$ using the rules:
        \State \vspace{-4mm}\begin{equation*}
    \label{Equation::LPM_rules}
    \begin{aligned}
        \text{If $\pi_i + \pi_j < 1$, then}\;(\pi_i, \pi_j) &= 
        \begin{cases} 
              (0, \pi_i + \pi_j) & \text{with probability}\; \frac{\pi_j}{\pi_i+\pi_j} \\
              (\pi_i + \pi_j, 0) & \text{with probability}\; \frac{\pi_i}{\pi_i+\pi_j}, 
        \end{cases}\\
        \text{else if $\pi_i + \pi_j \geq 1$, then}\;(\pi_i, \pi_j) &= 
        \begin{cases} 
              (1, \pi_i + \pi_j-1) & \text{with probability}\; \frac{1-\pi_j}{2-\pi_i-\pi_j} \\
              (\pi_i + \pi_j-1, 1) & \text{with probability}\; \frac{1-\pi_i}{2-\pi_i-\pi_j}. 
        \end{cases}
    \end{aligned}
\end{equation*}
    \State Set $\mathcal{U} = \mathcal{U} \setminus \{\datap_k \in \mathcal{U} : \pi_k\in\{0, 1\}\}$ \Comment{Remove samples with decision}
    \EndWhile
    \State Set $\dset = \{\datap_k \in \mathcal{U}^* : \pi_k = 1\}$ \Comment{Data points with positive sampling decision}
      \\
      \Return $\dset$ 
  \end{algorithmic}
  \label{Algorithm::LPM}
\end{algorithm}

This means that all the sampling decisions will be completed in at most $\lvert\mathcal{U}\rvert$ iterations of the algorithm. Note that in LPM, it is not required that $\pi_i = \pi_j\;\forall\, i,j$ but it is required that $\sum_{\datap_i\in \mathcal{U}} \pi_i = N$, where $N$ is the size of sampled data set $\dset$ \citep[see e.g.][]{Grafstrom2012}. The main steps of the LPM sampling are the following: 
\begin{enumerate}
    \item Randomly choose a data point $\datap_i\in \mathcal{U}$ with uniform probability.
    \item Find the nearest neighbor (i.e., nearest in e.g. Euclidean distance $e$ sense) $\datap_i$ of $\datap_j$ in the feature space $\mathcal{X}$.
    \item If data point $\datap_i$ has two neighbors equally close in the feature space, then randomly with equal probability select either of the two neighbors. 
    \item Update the inclusion probability pair $(\pi_i, \pi_j)$ using the rules found in Algorithm \ref{Algorithm::LPM}.
    \item Remove the data point in the pair $(\datap_i, \datap_j)$ for which the inclusion probability is either $0$ or $1$ from further consideration.
    \item If all the inclusion probabilities in set $\Pi $ have $\pi_k = 1$ or $\pi_k = 0$, then stop the algorithm and include data points with $\pi_k = 1$ into $\mathcal{D}$. Otherwise, repeat from step 1. 
\end{enumerate}
The corresponding pseudocode for LPM is shown in Algorithm \ref{Algorithm::LPM}.

\subsection{Data sampling via Bayesian optimization}\label{Section::BMVI}
The sampling method we propose is based on utilizing a prediction model's uncertainty on the value of response variable $y$ under a given input datum $\datax$. To give motivation for the proposed method, we note that typically we have observations of the response variable $y$ (such as forest growing stock) only in sampled points, whereas auxiliary data variables $\datax$ (e.g. satellite/airborne data) are often available throughout the entire inventory area. This may often be the case in inventories that use RS-based auxiliaries such as the Finnish multi-source NFI \citep{Tomppo2008}. We aim to utilize the relationship between the response variable and auxiliary information by firstly building a probabilistic model using the observed data set $\dset$, and then basing the sampling decision on the model's conditional uncertainty on the value of $y$ (quantified by its variance) given input feature datum $\datax$. A new sample point is to be chosen based on where the prediction model has the highest uncertainty on the value of $y$. Whereas the sampling decisions with methods like SRS or LPM focus mainly to variables $y$ and $\datax$ in itself, the sampling decisions in the proposed method are based on the functional relationship $y=f(\textbf{x}; \w)$. The proposed method thus assumes that there exists strong enough correlation between the predictor variables $\textbf{x}$ and the response variable $y$ in order to utilize this relationship in data sampling. This assumption is however necessary and fundamental to make in any data and modeling based analysis. We will next go through the proposed method in a more detailed manner. Most of the following is based on literature by e.g. \citep{BishopBook,MacKay1992InformationBasedOF,MacKay1992TheEF,MacKay1992BayesiInterp,Neal:1996}. Furthermore, more related literature based on Bayesian optimization can be found e.g. in the works of \citep{Xia2006,Chipeta2017,Muller2007,Zhu2006,Diggle2006,Rasmussen:2005:GPM:1162254,Snoek2012,Osborne2010,Werner2012}.    

Having observed a data set $\dset=\{(\datax_1, y_1), (\datax_2, y_2), ..., (\datax_N, y_N)\}$, we are interested in knowing the conditional distribution of $y$ given a new input vector $\xnew$ and the data set $\dset$. Explicitly put, we want to find out $p(y|\xnew, \dset)$, which can be written as: 
\begin{equation}
\label{Equation::posterior_predictive_y}
p(y|\xnew,\dset)=\int_{\mathbb{R}^q}p(y, \w|\xnew,\dset)\,d\w=\int_{\mathbb{R}^q}p(y|\xnew,\w)p(\w|\dset)\,d\w,    
\end{equation}
where $p(\w|\dset)$ is the posterior distribution of model parameters. Note also that $p(y|\xnew, \w)=p(y|\xnew, \w, \dset)$. This follows from the fact that the model parameters $\w$ and $\xnew$ completely determine the distribution of $y$ once the data $\dset$ has been observed. Using equation \ref{Equation::posterior_predictive_y}, we can now state the main statistic of interest in the proposed sampling method, which is the variance of the distribution $p(y|\datax, \dset)$, i.e.:
\begin{equation}
\label{Equation::posterior_pred_variance}
\yvar=E_y\left[\left(y-\mu\right)^2\rvert\datax, \dset\right],
\end{equation}
where $\mu$ is the mean value of $y$ w.r.t. distribution $p(y|\datax,\dset)$ and $E_y$ stands for expectation w.r.t. same distribution. We see from equation \ref{Equation::posterior_pred_variance} that the variance is a function of $\datax$, but not $\dset$ since we assume this to be fixed. We call the proposed sampling method (based on the statistic in equation  \ref{Equation::posterior_pred_variance}) \emph{Bayesian maximum variance inclusion} (BMVI). The BMVI always chooses sample data points $\datap = (\datax, y)$ where $\sigma_{\dset}^2(\datax)$ attains highest values (i.e. maximum uncertainty). The pseudocode for the BMVI is illustrated in Algorithm \ref{Algorithm::Bayesian_sampling}. The symbols $k, \dset_p, \dset_s$ denote the number of data points to be sampled, a prior data set available for calculating the posterior predictive distribution $p(y|\datax, \dset_p)$, and the new sampled data set (i.e. $k=\left|\dset_s\right|$). The algorithm shows that the inclusion probabilities are $\pi_i = 1$ for the $k$ single data samples with the highest posterior predictive variances. For all the remaining data points the inclusion probabilities are $\pi_i = 0$.

\begin{algorithm}[t]
  \caption{Pseudocode for BMVI}\label{Algo}
  \begin{algorithmic}[1]
    \Require $\dset_p, \mathcal{U}, k$ \Comment{Prior data set, sample population and sample size}
    \Ensure $\dset_{s}$ \Comment{Sample data set}
    \State Set $\dset_s = \emptyset$
    \State Calculate $p(y|\datax, \dset_p)$ using prior data set
        $\dset_p$ \Comment{Note $\dset_p\subset \mathcal{U}$}
      \For{$i \leftarrow 1$ to $k$} \Comment{Select $k$ data points}
        \State Set $(\datax_i, y_i) = \argmax_{(\datax, y) \in \mathcal{U}\setminus \dset_p} \yvar$ \Comment{Data point with max. uncertainty}
        \State Set $\dset_s = \dset_s \cup \{(\datax_i, y_i)\}$ \Comment{Include data point into sample}
        \State Set $\mathcal{U} = \mathcal{U} \setminus \{(\datax_i, y_i)\}$ \Comment{Remove sampled point from population}
      \EndFor
      \\
      \Return $\dset_{s}$ \Comment{Return sample of size $k$}
  \end{algorithmic}
  \label{Algorithm::Bayesian_sampling}
\end{algorithm}

After making Gaussian assumptions on the distributions in equation \ref{Equation::posterior_predictive_y}, it follows that the variance statistic of equation \ref{Equation::posterior_pred_variance} can be written as: 

\begin{equation}
\label{Equation::post_pred_variance_result_beg}
\sigma_{\dset}^2(\datax) = \frac{1}{\beta} + \g(\datax)^T\textbf{A}^{-1}\g(\datax),
\end{equation}

where $\beta > 0$ is a parameter controlling the prior variance of the response variable $y$, $\g$ is a gradient vector of the prediction model $f(\datax; \w)$ evaluated at a maximum posterior point, and $\textbf{A}$ is the Hessian matrix of the exponent of the posterior distribution of model weights $\w$. Detailed definitions and derivations of this result can be found from the appendix part of this study. A Python implementation and example demonstration of the BMVI method can be found from \citep{BMVI_GITHUB}.

\subsection{Prediction models}
Next, we will give a short introduction to the prediction models $f(\textbf{x}; \w)$ we apply in the empirical analyses in section \ref{Section:analysis_and_results}.
\subsubsection{Ridge regression}
The first prediction method used in our analyses is ridge regression known also as regularized least squares, RLS \citep[see e.g.][]{Bishop::Pattern_recog_ML_2006}. RLS is almost identical to basic linear regression method, with the exception that instead of minimizing simply the squared error between observed data and predictions, the RLS adds a regularizing term into the squared error minimization. This addition makes the model selection process to favor more well-behaving models, which are more likely to achieve successful generalization to new unseen data \citep[see e.g.][]{Vapnik}. Explicitly, in RLS the prediction model is simply a linear function of the input data, i.e. $f(\datax; \w) = \w^T\datax + \theta_0$ where $\theta_0$ denotes the constant bias term of the model. In RLS, the model parameters are selected so as to minimize the (error) function:  
\begin{equation}
\label{Equation::S(w)_definitions_RLS}
    S(\w)=\frac{\beta}{2}\sum_{i=1}^N \{y_i-\w^T\datax_i-\theta_0\}^2+\frac{\alpha}{2}\sum_{j=1}^m \theta_j^2,
\end{equation}
where $\alpha, \beta > 0$ and $\alpha$ controls the degree of regularization. The constants $\alpha, \beta$ correspond directly to those in equations \ref{Equation::w_prior}, \ref{Equation::y_prior_distribution} and \ref{Equation::w_posterior}, showing the connection between Tikhonov regularization and Bayesian modeling \citep[see e.g.][]{Murphy:2012:MLP:2380985}. Note that it is not necessary to include the constant term $\theta_0$ into the second term in equation \ref{Equation::S(w)_definitions_RLS} since it simply controls the offset of the hyperplane $f(\datax; \w)$ but not its slopes. In our analyses, the RLS hyperparameter selection (i.e. $\alpha, \beta$) was conducted using leave-one-out cross-validation \citep[LOOCV,][]{gelman2013bayesian}.   

\subsubsection{Multilayer perceptron}
In addition to the RLS, a multilayer perceptron \citep[MLP,][]{BishopBook} was tested as a prediction model. A MLP is a feedforward neural network defined by the number of hidden layers $L$, inputs and outputs, hidden nodes and types of activation functions, and it has shown great performance in a number of applications. The MLP network is trained by minimizing a suitable error function, such as $S(\w)$ in the equation \ref{Equation::S(w)_definitions_RLS}. The parameters of a MLP can be defined as the set:
\begin{equation}
    \w \equiv \left\{\theta_{ij}^{(l)} \mid 1\leq l \leq L+1, 0 \leq i \leq d^{(l-1)}, 1 \leq j \leq d^{(l)} \right\},
\end{equation}
where $d^{(l)}$ is the number of nodes on layer $l$. In other words, $\theta_{ij}^{(l)}$ means a network weight connecting node $i$ at layer $l-1$ to node $j$ at layer $l$. The weights $\theta^{(1)}_{ij}$ and $\theta^{(L+1)}_{ij}$ correspond to weights connected to the input and output nodes respectively. As an example, a MLP with one hidden layer ($L=1$) can be explicitly expressed as a function: 
\begin{equation}
    f(\datax; \w) = f_2\left(\sum_{j=1}^{d^{(1)}} \theta_{j1}^{(2)} f_1\left(\sum_{i=1}^m \theta_{ij}^{(1)}x_i\right)\right),
\end{equation}
where now weights $\theta_{ij}^{(1)}$ and $\theta_{j1}^{(2)}$ correspond to connections of the hidden layer to input and output layers correspondingly. The functions $f_1(\cdot)$ and $f_2(\cdot)$ correspond to the activation functions, which need not be the same at all layers. Common choices for the activation functions are e.g. linear or sigmoid functions. In our experiments, we used a MLP model provided by the NETLAB-library \citep{Nabney2004}. The MLP network was trained using the scaled conjugate gradient algorithm \citep{Bazaraa:2013:NPT:2553227}.

\subsection{Implementation details of the empirical analysis}\label{Section::implementation_details}

\begin{figure}[b!]
\centering
\subfigure[]{\includegraphics[width=.49\textwidth]{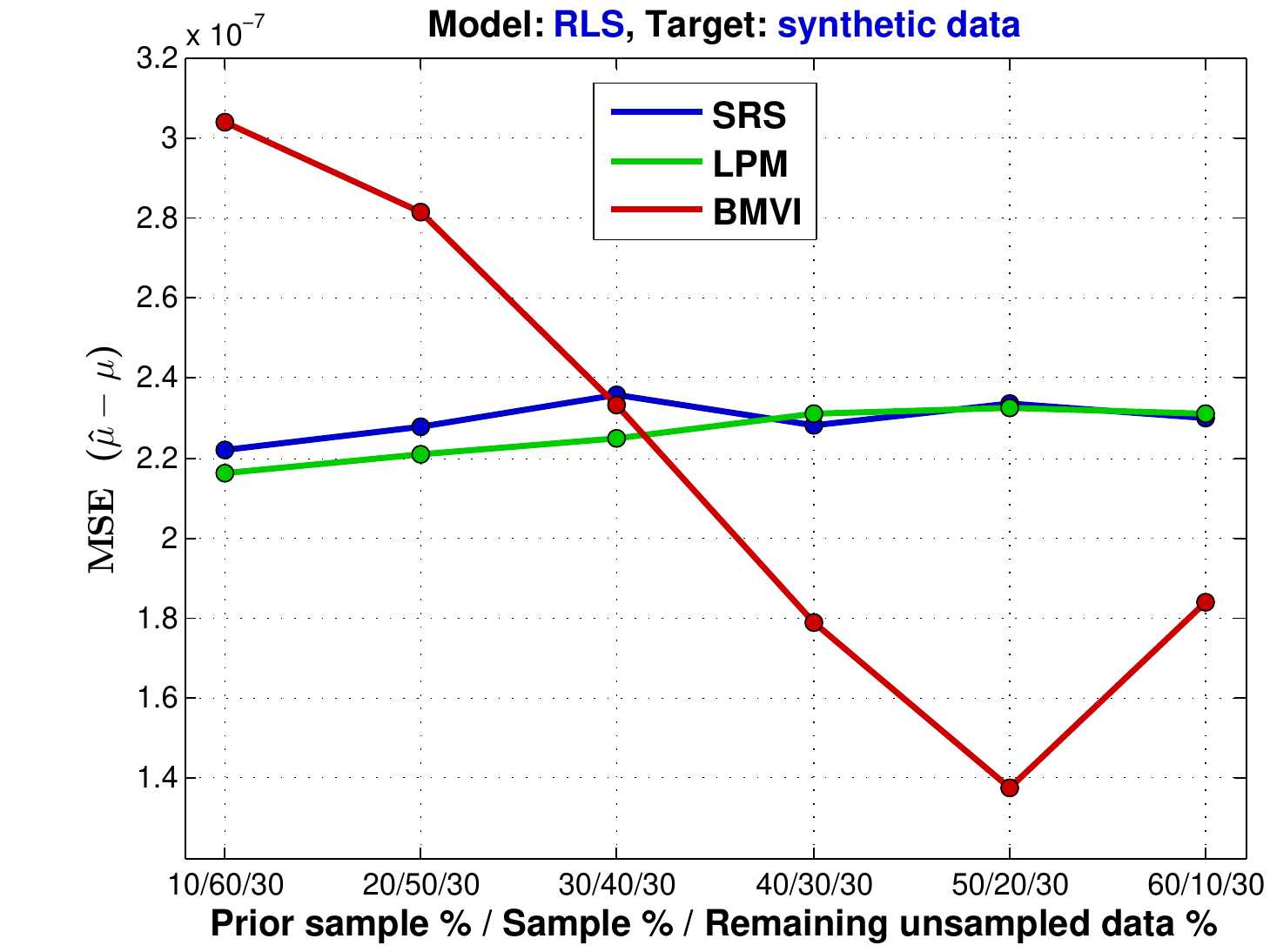}}%
\hspace{1mm}
\subfigure[]{\includegraphics[width=.49\textwidth]{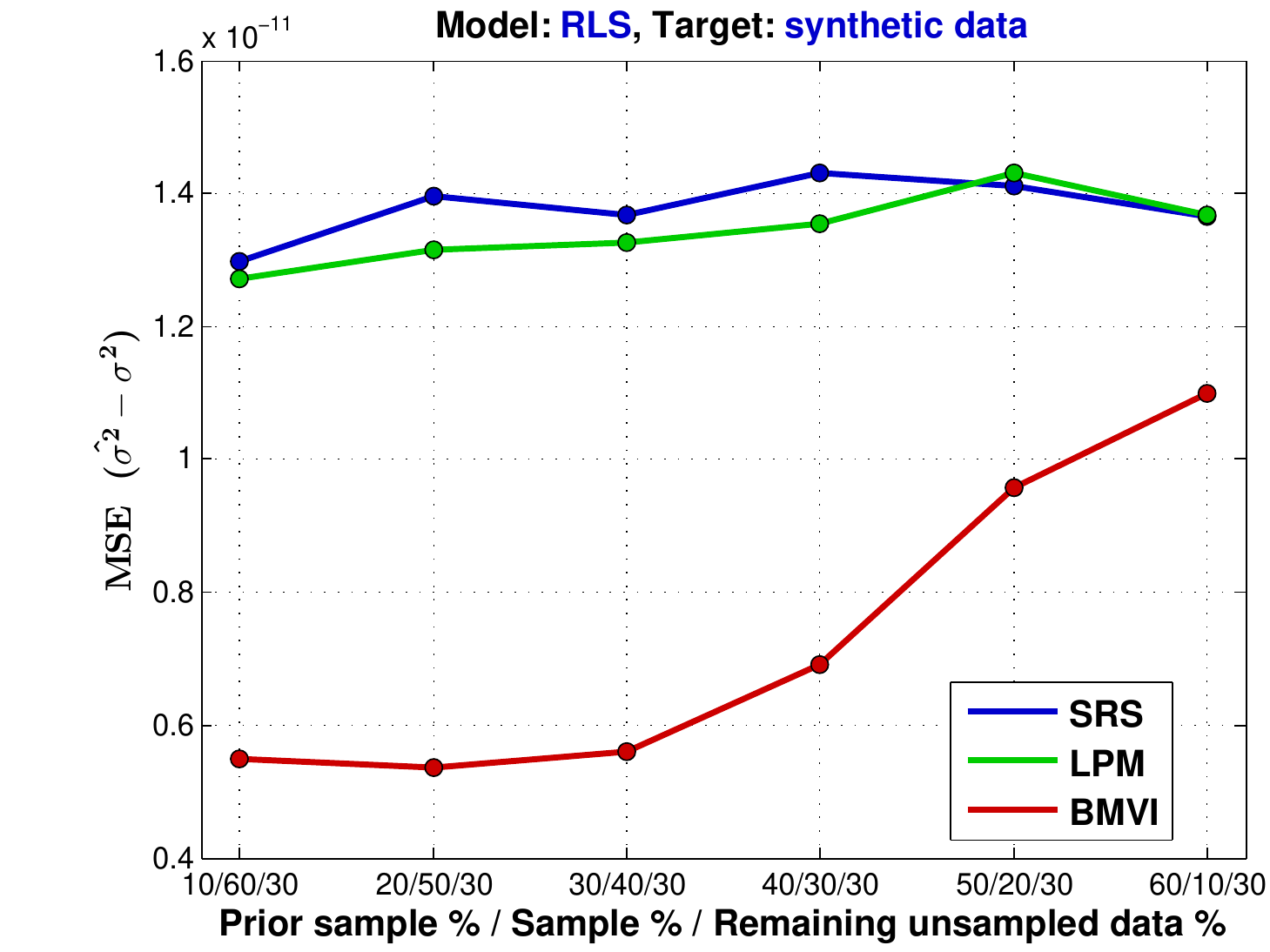}}%

\subfigure[]{\includegraphics[width=.49\textwidth]{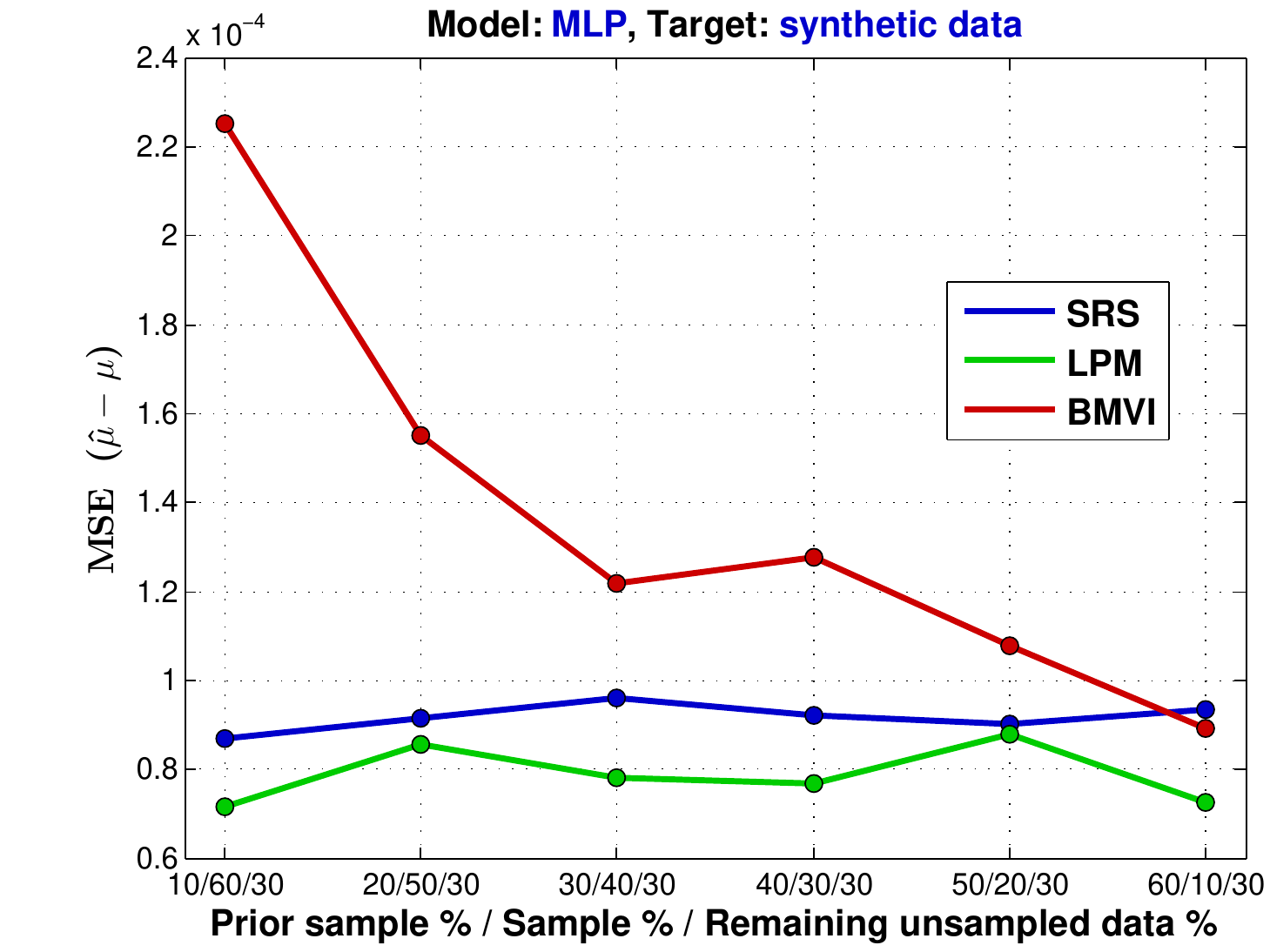}}%
\hspace{1mm}
\subfigure[]{\includegraphics[width=.49\textwidth]{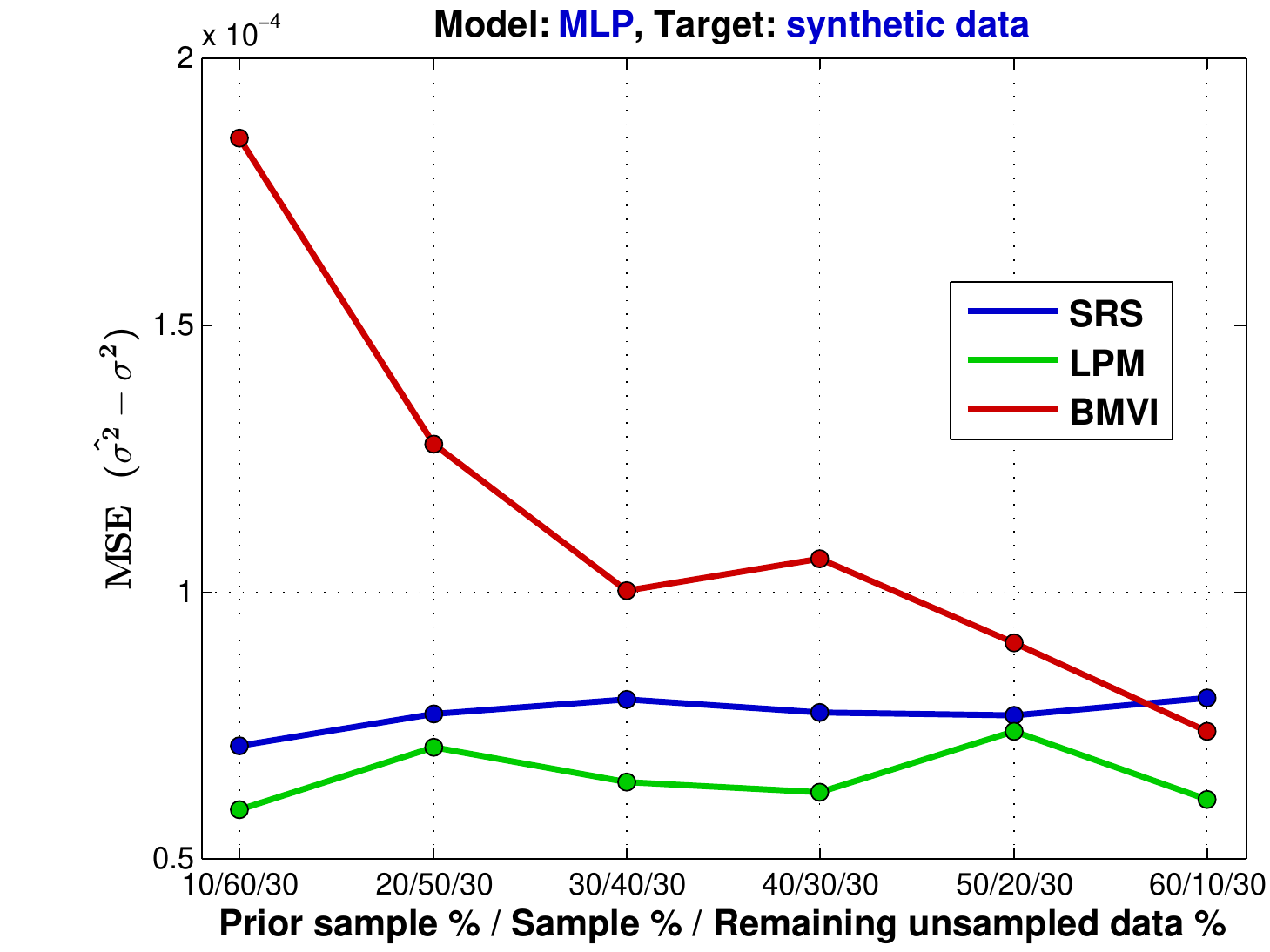}}%

\caption{Results of the empirical comparison of the \BM against SRS and LPM for the synthetic GMM data. (a)-(b): Population mean and variance estimation performance plots for RLS prediction model. The x-axis represents different values in the fraction vector $\fracset$ and y-axis represents the mean squared error value between estimated and true population parameters. (c)-(d): Analogous results as in (a)-(b) but for MLP prediction model.}
\label{Figure::Analysis_results_synthetic}
\end{figure}

Lastly, in this section we will describe the technical details of the empirical analyses in order to make it more clear on how to interpret the results in section \ref{Section:analysis_and_results}. The results of section \ref{Section:analysis_and_results} (Figures \ref{Figure::Analysis_results_synthetic}, \ref{Figure::Analysis_results_volall}, \ref{Figure::Analysis_results_volpine}, \ref{Figure::Analysis_results_volspruce}, \ref{Figure::Analysis_results_volbroadleaf} and Tables \ref{Table::results_mean} and \ref{Table::results_variance}) were produced using the algorithm presented in this section. Note that the emphasis of this study was not to find an optimal prediction model (like e.g. the RLS or MLP) for the data sets, but the comparison of the sampling methods by their performance in the estimation of response variable population parameters. Thus due to their irrelevance, no optimal prediction model parameters are listed in this study. Recall, that we denoted the data population as $\mathcal{U}$ and the prediction model as $f$. In addition, we will denote a sampling method as $S_M$, i.e. $S_M \in \{\text{SRS, LPM, BMVI}\}$, and sample data sets as $\dset_p, \dset_s \subset \mathcal{U}$ where we have $\dset_p \cap \dset_s = \emptyset$. 

Since the core principle behind the \BM sampling method is in utilizing the learned functional relationship between $\datax$ and $y$, the method assumes that we have some prior data set $\dset_p$ available for training the model $f$ before we conduct the sampling of new data, i.e. $\dset_s$ via BMVI. In clearer terms, the \BM uses previously sampled data to optimize future sampling decisions. Thus in the empirical experiments, it is always assumed that we have some prior data set $\dset_p$ available before the actual sampling of $\dset_s$ is conducted with given $S_M$. Furthermore, since the sampling decisions of the \BM method are obviously affected by the data used for training the prediction model $f$, it is of interest to study how the size of the prior training data $\dset_p$ with respect to the whole data population $\mathcal{U}$ affects the sampling performance of the BMVI. For this reason, we parameterize our experiments with a vector:

\begin{equation}
\label{Equation::Fraction_set}
\fracset = \left(\frac{\left|\dset_p\right|}{\left|\mathcal{U}\right|}, \frac{ \left|\dset_s\right|}{\left|\mathcal{U}\right|}, \frac{\left|\mathcal{U}\setminus (\dset_p\cup\dset_s)\right|}{\left|\mathcal{U}\right|} \right)\in (0,1)^3.
\end{equation}

In other words, the elements of the vector $\fracset$ are interpreted as: 1) the fraction of data points of the population $\mathcal{U}$ available in the prior set $\dset_p$, 2) the fraction of new data to be sampled into set $\dset_s$ with a given sampling method $S_M$, and 3) the remaining fraction of the population data (i.e. out-of-sample data) used for testing the estimation performance of population parameters. In our experiments we used a reasonable constant fraction of 30\% of the data for testing the estimation performance. Thus we always had in the experiments that $\frac{\left|\dset_p\right|}{\left|\mathcal{U}\right|}+\frac{\left|\dset_s\right|}{\left|\mathcal{U}\right|} = 0.7$, with $\frac{\left|\dset_p\right|}{\left|\mathcal{U}\right|},\frac{\left|\dset_s\right|}{\left|\mathcal{U}\right|}\in \{0.1, 0.2, 0.3, 0.4, 0.5, 0.6\}$.

The complete procedure used for obtaining the results of section \ref{Section:analysis_and_results} is described in Algorithm \ref{Algorithm::analysis_explained}. The algorithm is parametrized by the used data set $\mathcal{U}$, a fraction vector $\fracset$ and a prediction model $f$. The algorithm returns for all three sampling methods $S_M$ the mean squared error (MSE) values of the population mean $\mu$ and variance $\sigma^2$ parameter estimations. Note on line 3 of the algorithm that we repeat the experiments 100 times. This is due to decrease the effect of randomness in the estimation statistics by providing averaged results. For guaranteeing a valid comparison in the analysis results, all the sampling methods (i.e. SRS, LPM, BMVI) shared the same prior data set $\dset_p$ when implementing a single comparative calculation run (line 4 in the algorithm). Also, note in line 6 that only the BMVI method is dependent on $f$ and $\dset_p$. Rest of the algorithm is straightforward and on lines 10-11 the population parameters are estimated using the data set $\dset = \dset_p \cup \dset_s$ and the auxiliary $\datax$ data available in the set $V$. Recall that all the auxiliary data $\datax$ (i.e. RS data) is assumed to be fully known throughout the research area and the response variable data $y$ (e.g. tree volume) is only partly known and requires further sampling.

\setcounter{algorithm}{2}
\begin{algorithm}[t]
  \caption{Procedure used for obtaining the empirical results of section \ref{Section:analysis_and_results}}
  \begin{algorithmic}[1]
    \Require $\mathcal{U}, \fracset, f$ \Comment{Population data, sample fraction vector and prediction model}
    \Ensure $MSE_{\mu}^{\text{SRS}}, MSE_{\sigma^2}^{\text{SRS}}, MSE_{\mu}^{\text{LPM}}, MSE_{\sigma^2}^{\text{LPM}}, MSE_{\mu}^{\text{BMVI}}, MSE_{\sigma^2}^{\text{BMVI}}$
    \State Set $SE_{\mu}^{\text{SRS}} = \emptyset,\; SE_{\mu}^{\text{LPM}} = \emptyset,\; SE_{\mu}^{\text{BMVI}} = \emptyset$ \Comment{Sets of squared error values}
    \State Set $SE_{\sigma^2}^{\text{SRS}} = \emptyset,\; SE_{\sigma^2}^{\text{LPM}} = \emptyset,\; SE_{\sigma^2}^{\text{BMVI}} = \emptyset$ 
      \For{$i \leftarrow 1$ to $100$} \Comment{Repeat 100 times to produce averaged results}
        \State Select a random prior sample set $\mathcal{D}_{p}$ from $\mathcal{U}$ according to $\fracset$
        \For{$S_M\in \{\text{SRS, LPM, BMVI}\}$} \Comment{Do sampling with all methods}
        \State Select a sample $\mathcal{D}_s$ from $\mathcal{U} \setminus \mathcal{D}_{p}$ using $S_M, \fracset, f$ and $\mathcal{D}_{p}$
        \State Set $\dset = \mathcal{D}_{p} \cup \mathcal{D}_{s}$ \Comment{Combine prior and sampled data}
        \State Set $V = \mathcal{U} \setminus \dset$ \Comment{Use the remaining unsampled data for testing}
        \State Train a prediction model $f$ using data set $\dset$
        \State Set estimator $\hat{\mu} = \left|\mathcal{U}\right|^{-1}\left(\sum_{\textbf{d}\in \dset} y + \sum_{\textbf{d}\in V} f(\textbf{x})\right)$
        \State Set estimator $\hat{\sigma^2} = (\left|\mathcal{U}\right|-1)^{-1}\left(\sum_{\textbf{d}\in \dset} (y-\hat{\mu})^2 + \sum_{\textbf{d}\in V} (f(\textbf{x})-\hat{\mu})^2\right)$
        \State Set $SE_{\mu}^{S_M}[i] = (\hat{\mu}-\mu)^2$ \Comment{Error between estimate and true value}
        \State Set $SE_{\sigma^2}^{S_M}[i] = (\hat{\sigma^2}-\sigma^2)^2$
        \EndFor
      \EndFor
      \For{$S_M\in \{\text{SRS, LPM, BMVI}\}$} \Comment{Calculate MSEs for all methods}
      \State Set $MSE_{\mu}^{S_M} = \text{mean}\left(SE_{\mu}^{S_M}\right)$
      \State Set $MSE_{\sigma^2}^{S_M} = \text{mean}\left(SE_{\sigma^2}^{S_M}\right)$
      \EndFor \Comment{Lastly return all MSE values for all methods}
      \\
      \Return $MSE_{\mu}^{\text{SRS}}, MSE_{\sigma^2}^{\text{SRS}}, MSE_{\mu}^{\text{LPM}}, MSE_{\sigma^2}^{\text{LPM}}, MSE_{\mu}^{\text{BMVI}}, MSE_{\sigma^2}^{\text{BMVI}}$ 
  \end{algorithmic}
  \label{Algorithm::analysis_explained}
\end{algorithm}

The auxiliary predictor features used in the real world data case are listed in Table \ref{table:case_data_sets}. The response variables for synthetic GMM and volume of growing stock (all trees, pine trees, spruce trees, broadleaf trees) are denoted in the results in Tables \ref{Table::results_mean} and \ref{Table::results_variance} as synt, $v_a, v_p, v_s$ and $v_b$ respectively.

\section{Results}\label{Section:analysis_and_results}
In this section, we will go through the empirical results of comparing the \BM method with SRS and LPM sampling using the data sets described in section \ref{Section::Data_and_research_area}. Refer to Algorithm \ref{Algorithm::analysis_explained} in section \ref{Section::implementation_details} for technical details on the results.

\subsection{Case 1: synthetic data}
In Figure \ref{Figure::Analysis_results_synthetic} we see the results of population parameter estimation for the synthetically generated GMM data set with RLS and MLP prediction models. We can notice from Figure \ref{Figure::Analysis_results_synthetic} (a) that the BMVI eventually achieves the best performance from the three sampling methods when a RLS prediction model is used. It is expected that the performance of the BMVI improves as the amount of data in set $\dset_p$ increases prior to sampling new data $\dset_s$. This is clear because the probability of successful estimation of the functional relationship $f$ between auxiliary and response variable increases as more data becomes available for model training. Interestingly, in the case of variance estimation in Figure \ref{Figure::Analysis_results_synthetic} (b) the BMVI always achieves best results, although the MSE value shows an increasing trend when more data becomes available prior to sampling. By investigating the synthetic data in Figure \ref{Figure::synthetic_data}, it is in part explained why the BMVI always achieves best performance in population variance estimation. Most of the synthetic data variation occurs in the edges of the GMM data, which is exactly where the BMVI samples data in the case of a linear RLS model. In Figures \ref{Figure::Analysis_results_synthetic} (c)-(d) we notice the analogous results as before but for MLP prediction model. In this case, we notice the BMVI does not outperform SRS or LPM, regardless of the size of the prior data $\dset_p$. 

The results show, although artificial, that the prediction model $f$ has a significant effect on the performance of the BMVI method. Thus, the results suggest that it is important to have well-founded justification for using a given prediction model family with the BMVI method, if we are to expect the BMVI to outperform SRS and LPM in population parameter estimations. 

\begin{figure}[b!]
\centering
\subfigure[]{\includegraphics[width=.47\textwidth]{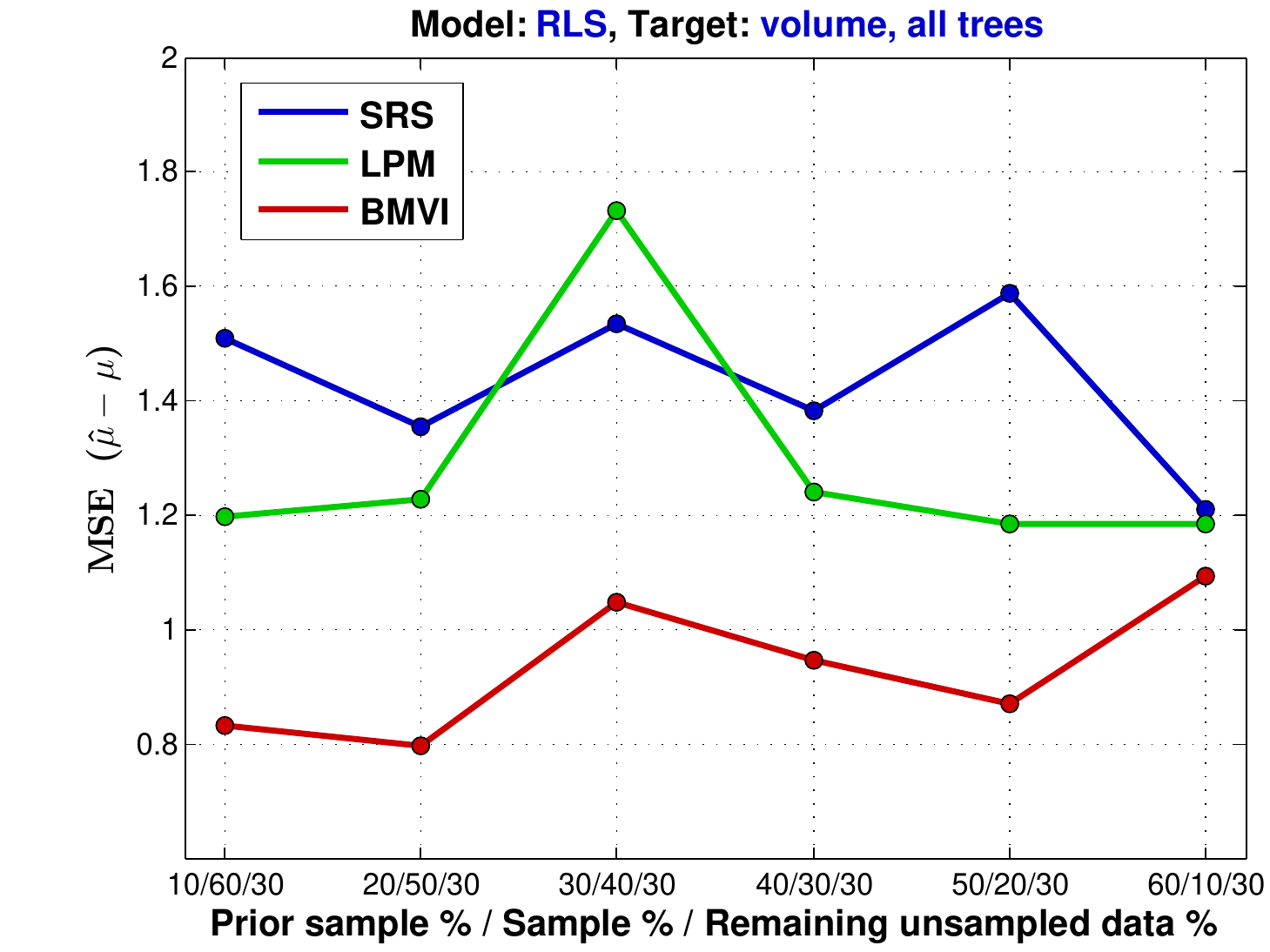}}%
\hspace{1mm}
\subfigure[]{\includegraphics[width=.47\textwidth]{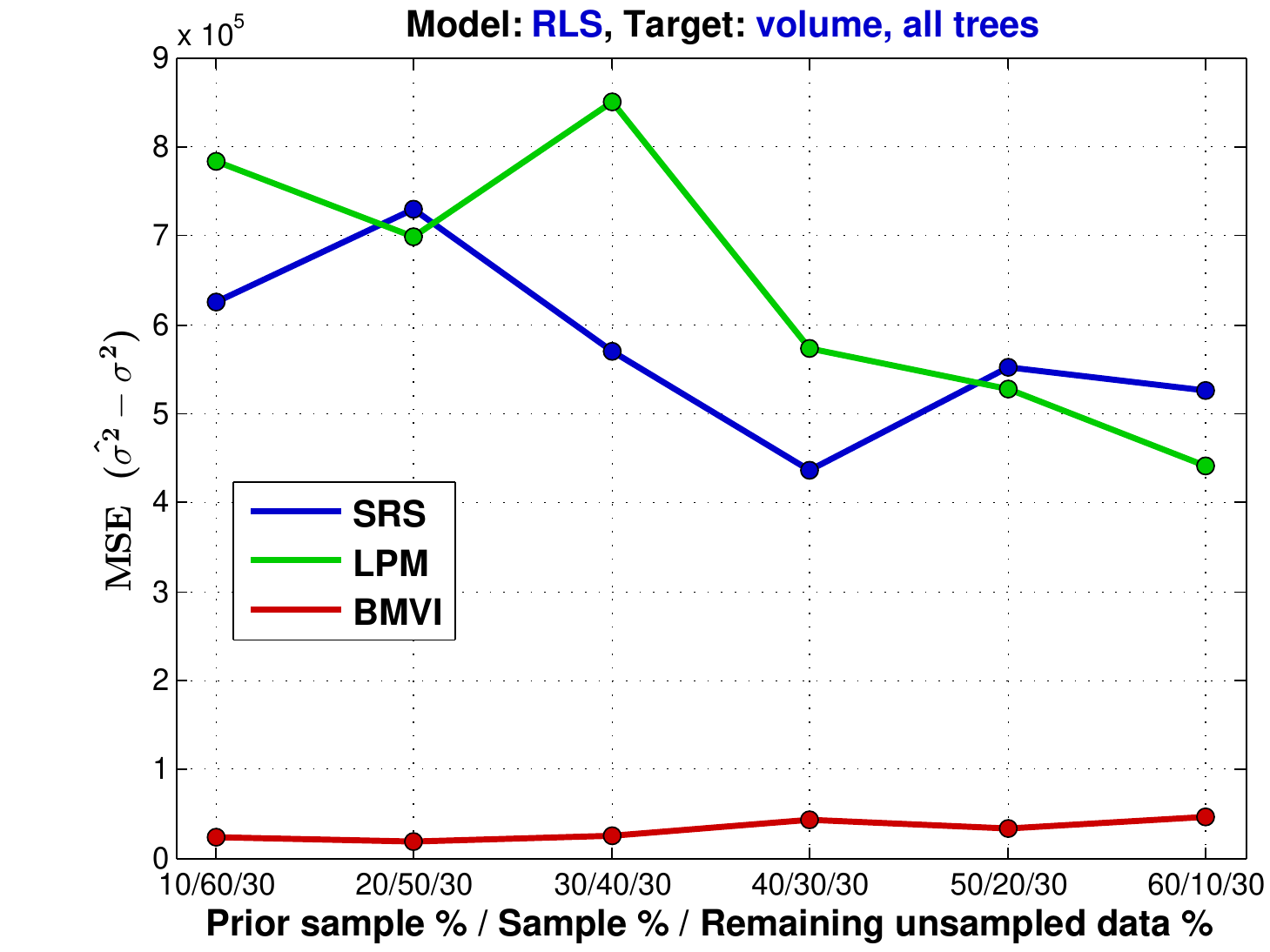}}%

\subfigure[]{\includegraphics[width=.47\textwidth]{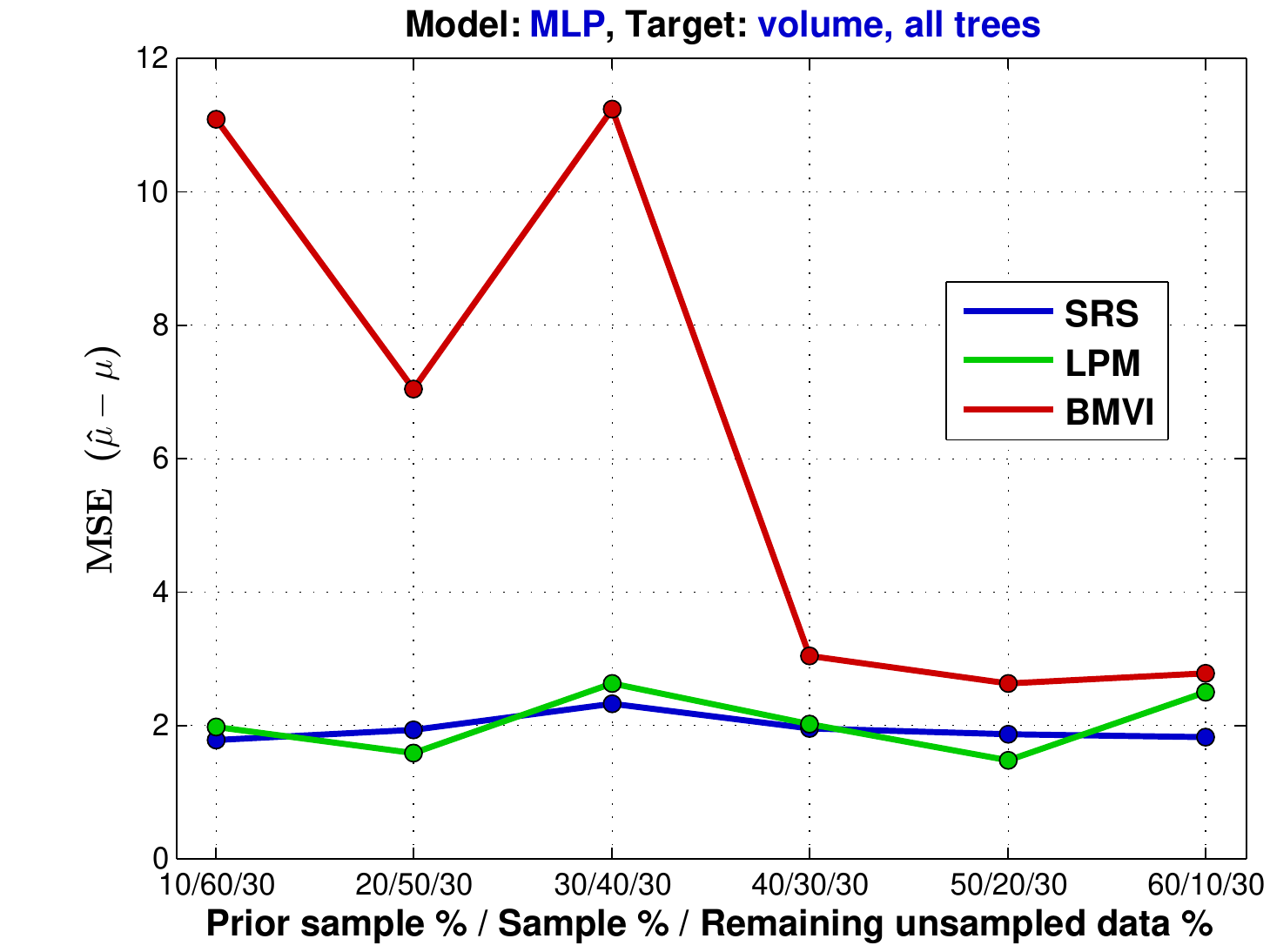}}%
\hspace{1mm}
\subfigure[]{\includegraphics[width=.47\textwidth]{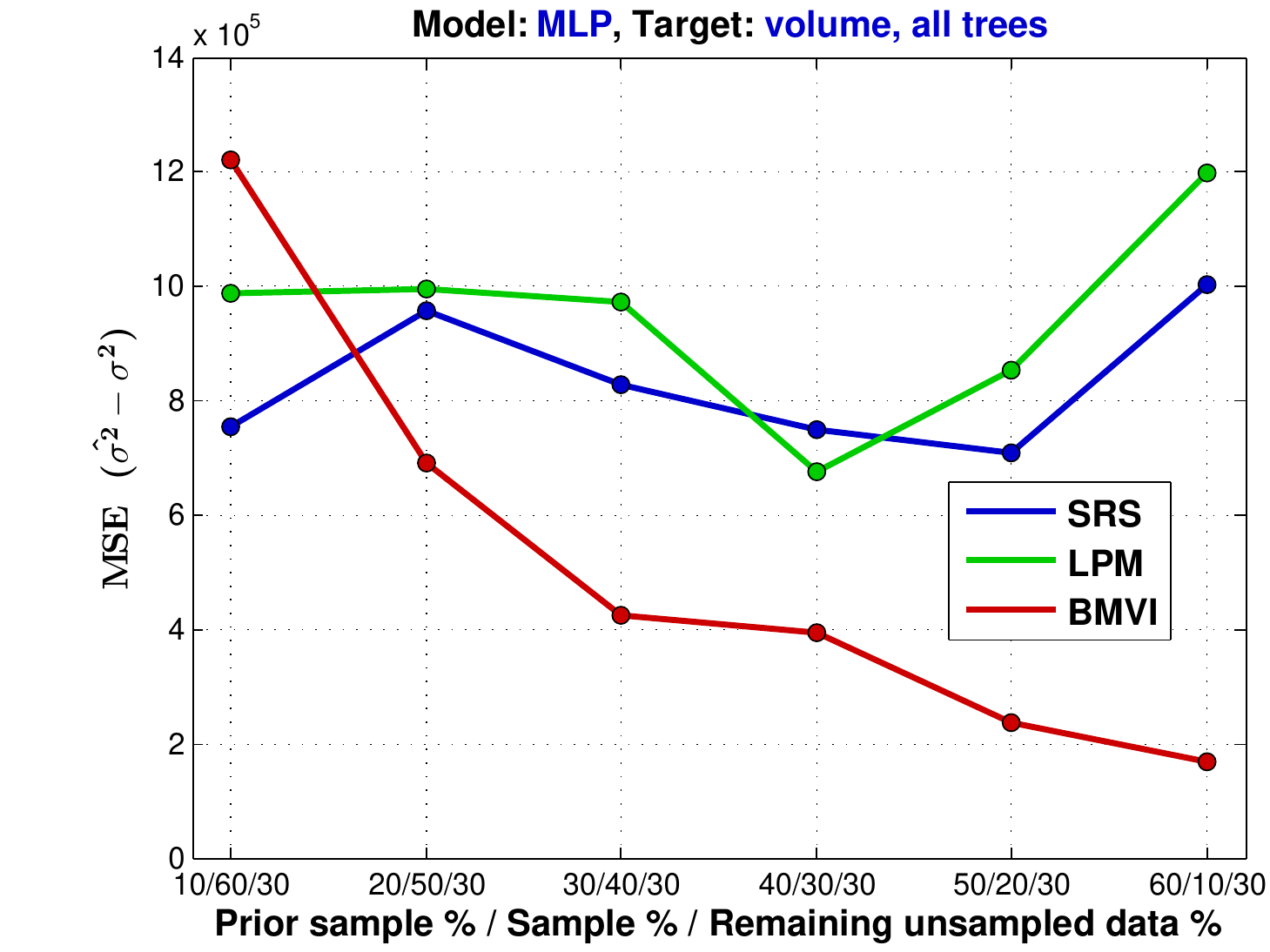}}%

\caption{Results of the empirical comparison of the \BM against SRS and LPM for the response variable: total volume of growing stock, all trees. (a)-(b): Population mean and variance estimation performance plots for RLS prediction model. The x-axis represents different values in the fraction vector $\fracset$ and y-axis represents the mean squared error value between estimated and true population parameters. (c)-(d): Analogous results as in (a)-(b) but for MLP prediction model.}
\label{Figure::Analysis_results_volall}
\end{figure}

\begin{figure}[b!]
\centering
\subfigure[]{\includegraphics[width=.47\textwidth]{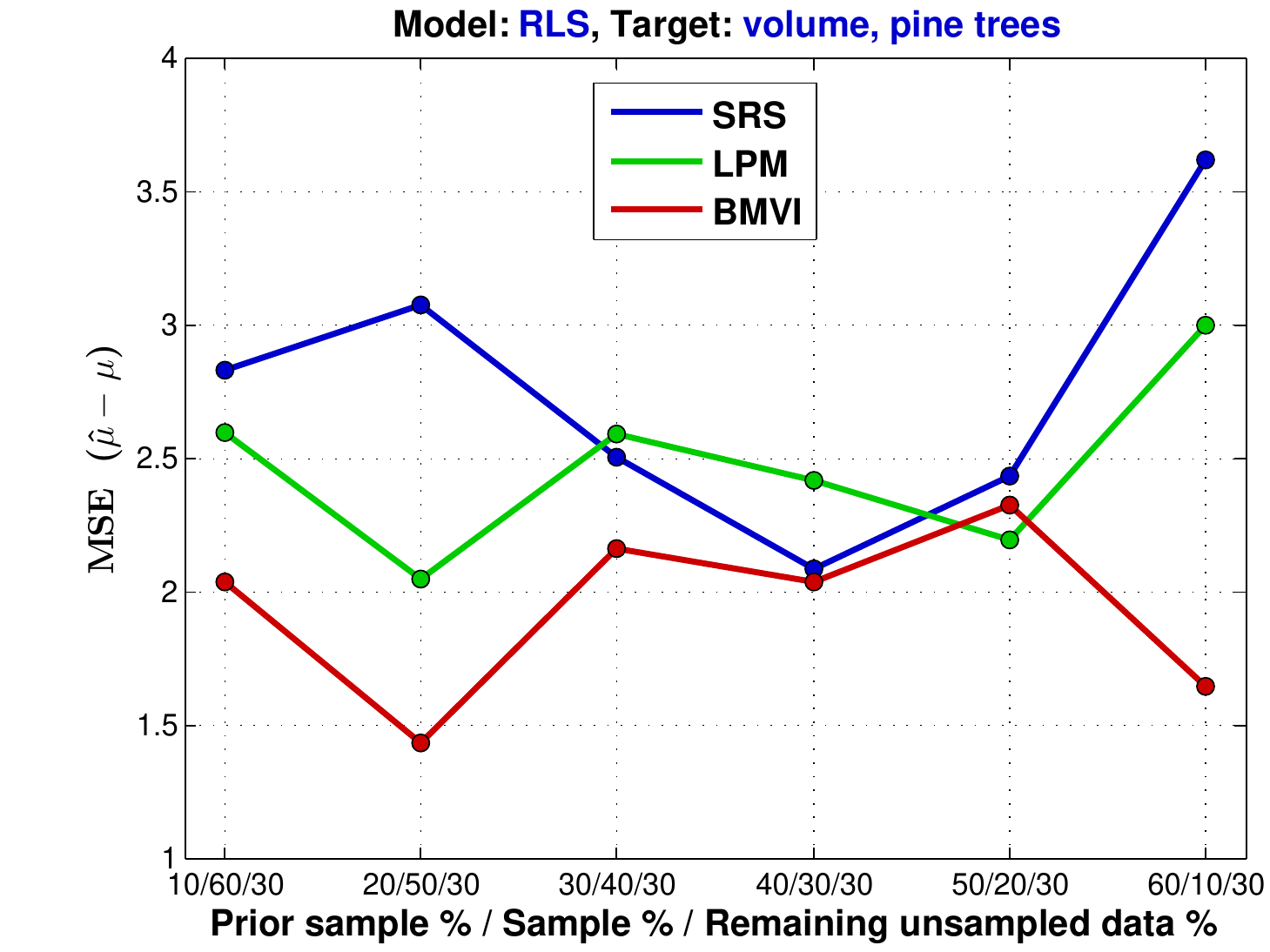}}%
\hspace{1mm}
\subfigure[]{\includegraphics[width=.47\textwidth]{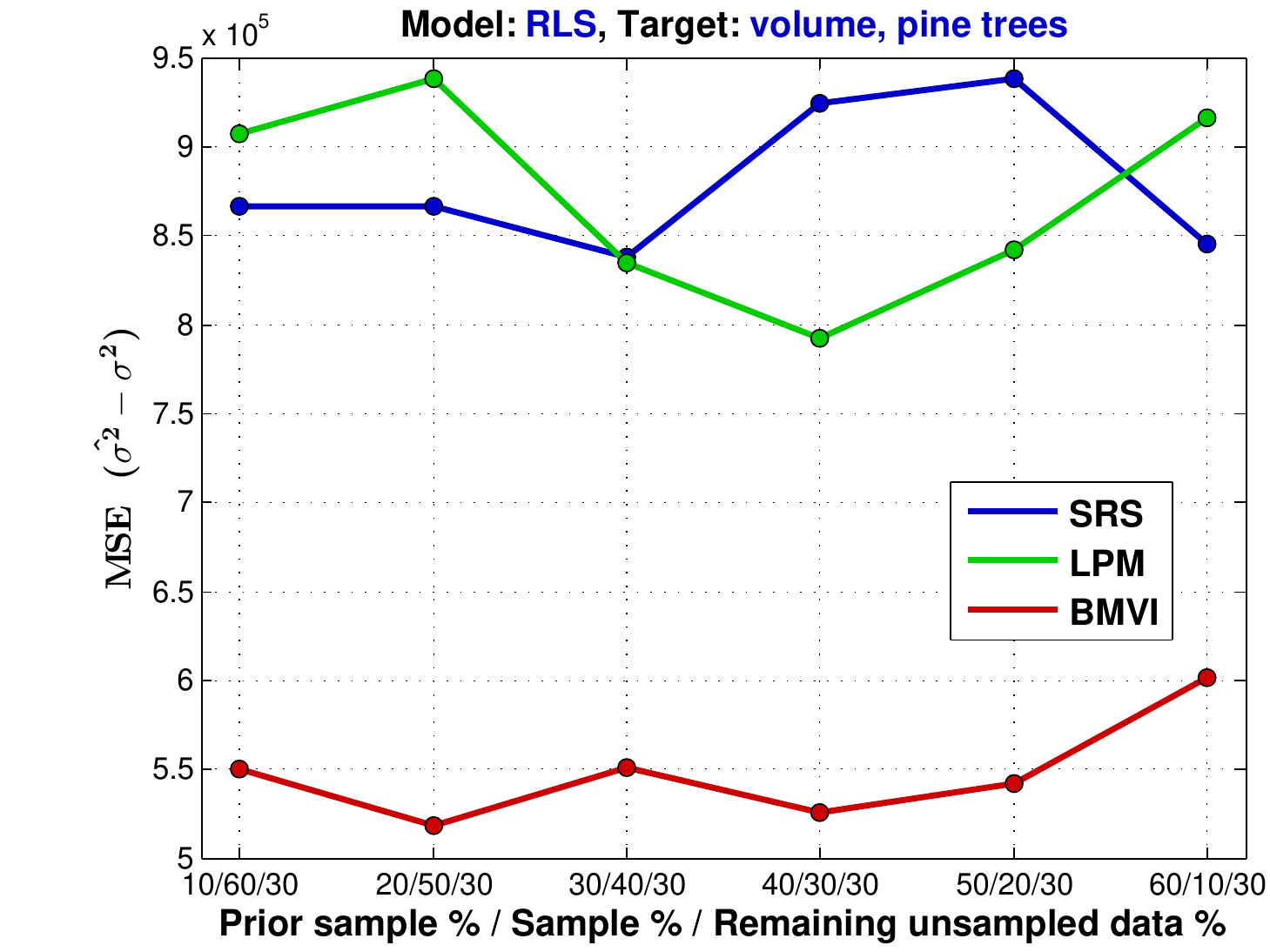}}%

\subfigure[]{\includegraphics[width=.47\textwidth]{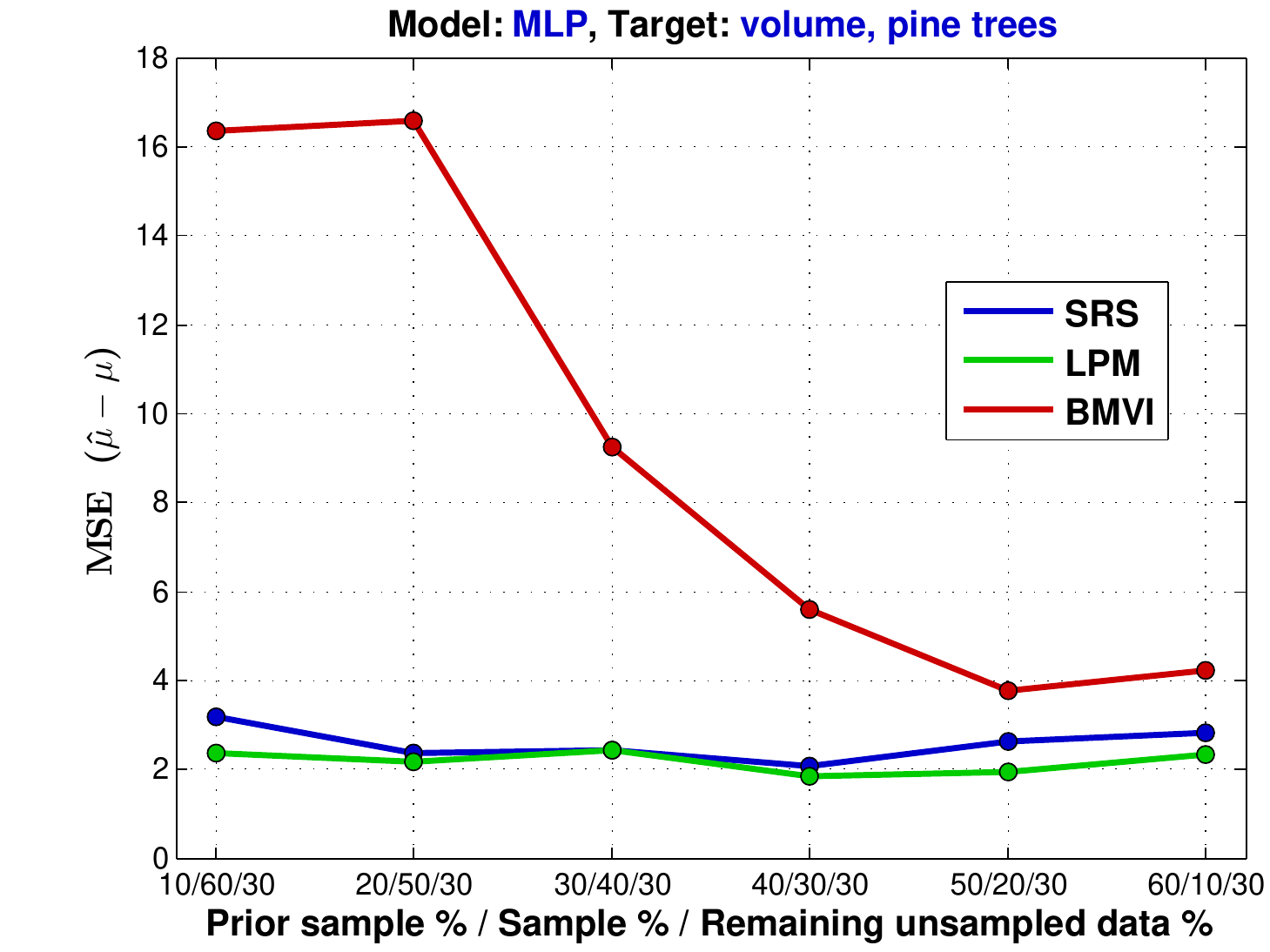}}%
\hspace{1mm}
\subfigure[]{\includegraphics[width=.47\textwidth]{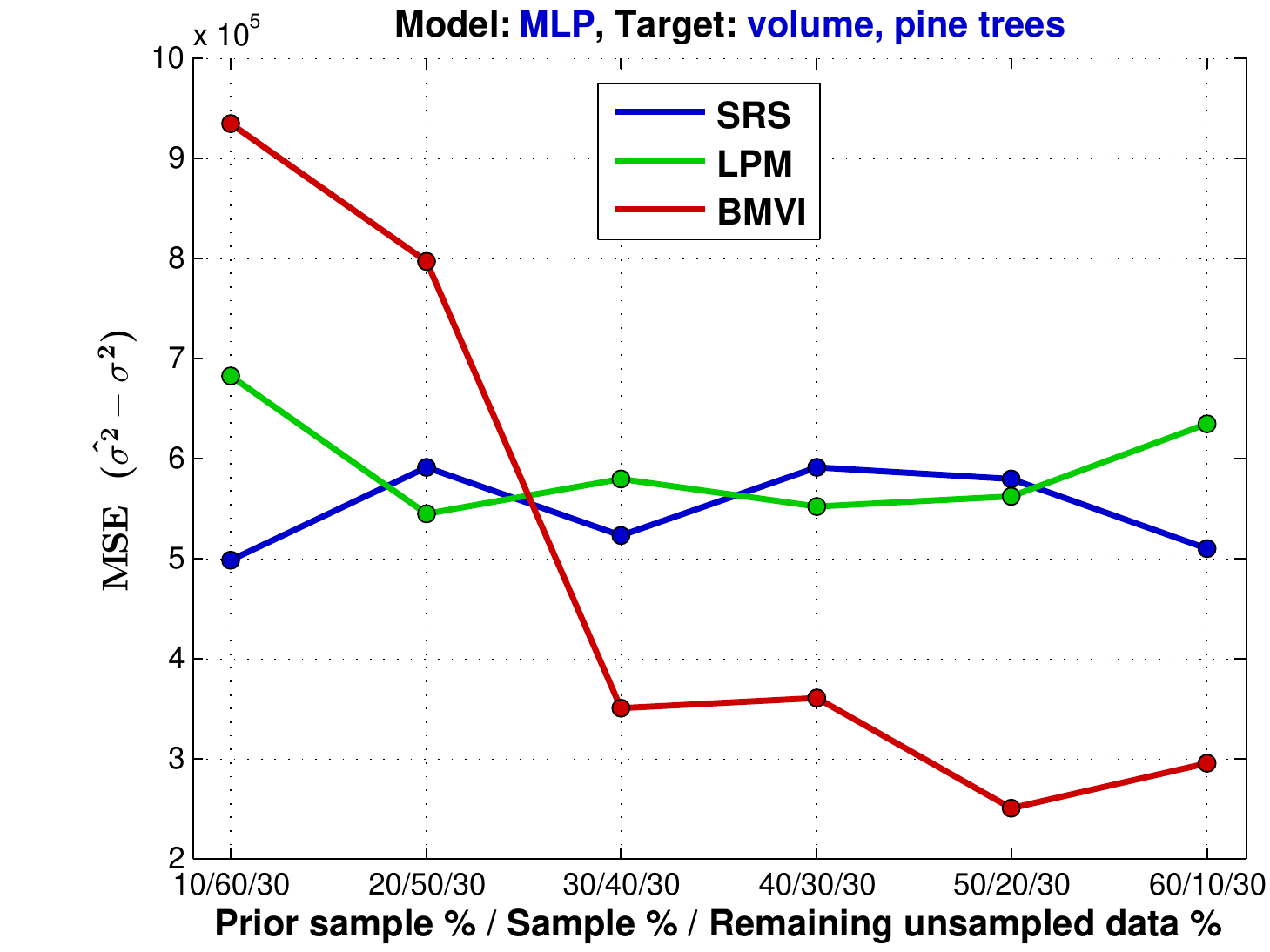}}%

\caption{Results of the empirical comparison of the \BM against SRS and LPM for the response variable: total volume of growing stock, pine trees. (a)-(b): Population mean and variance estimation performance plots for RLS prediction model. The x-axis represents different values in the fraction vector $\fracset$ and y-axis represents the mean squared error value between estimated and true population parameters. (c)-(d): Analogous results as in (a)-(b) but for MLP prediction model.}
\label{Figure::Analysis_results_volpine}
\end{figure}

\begin{figure}[b!]
\centering
\subfigure[]{\includegraphics[width=.47\textwidth]{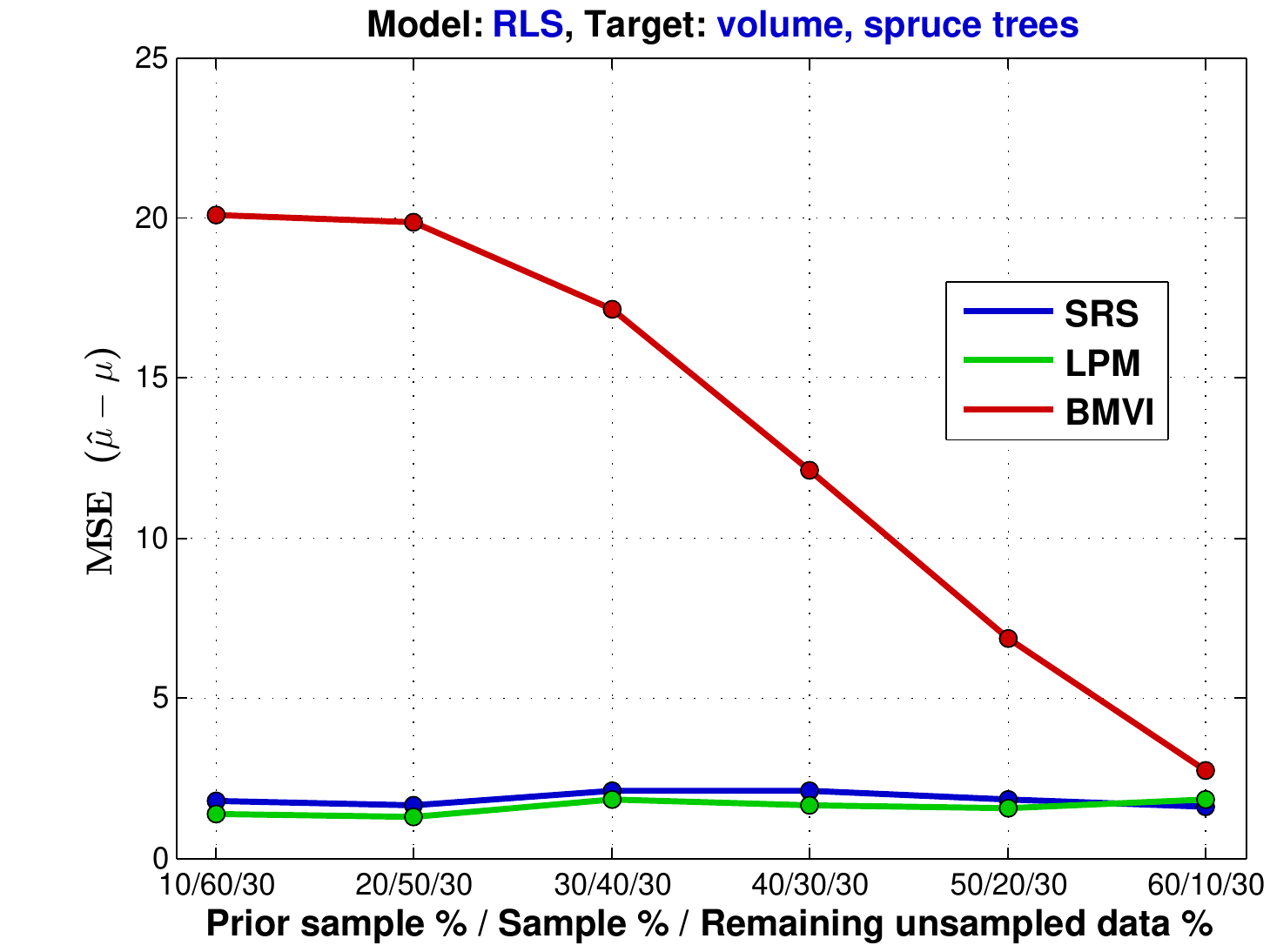}}%
\hspace{1mm}
\subfigure[]{\includegraphics[width=.47\textwidth]{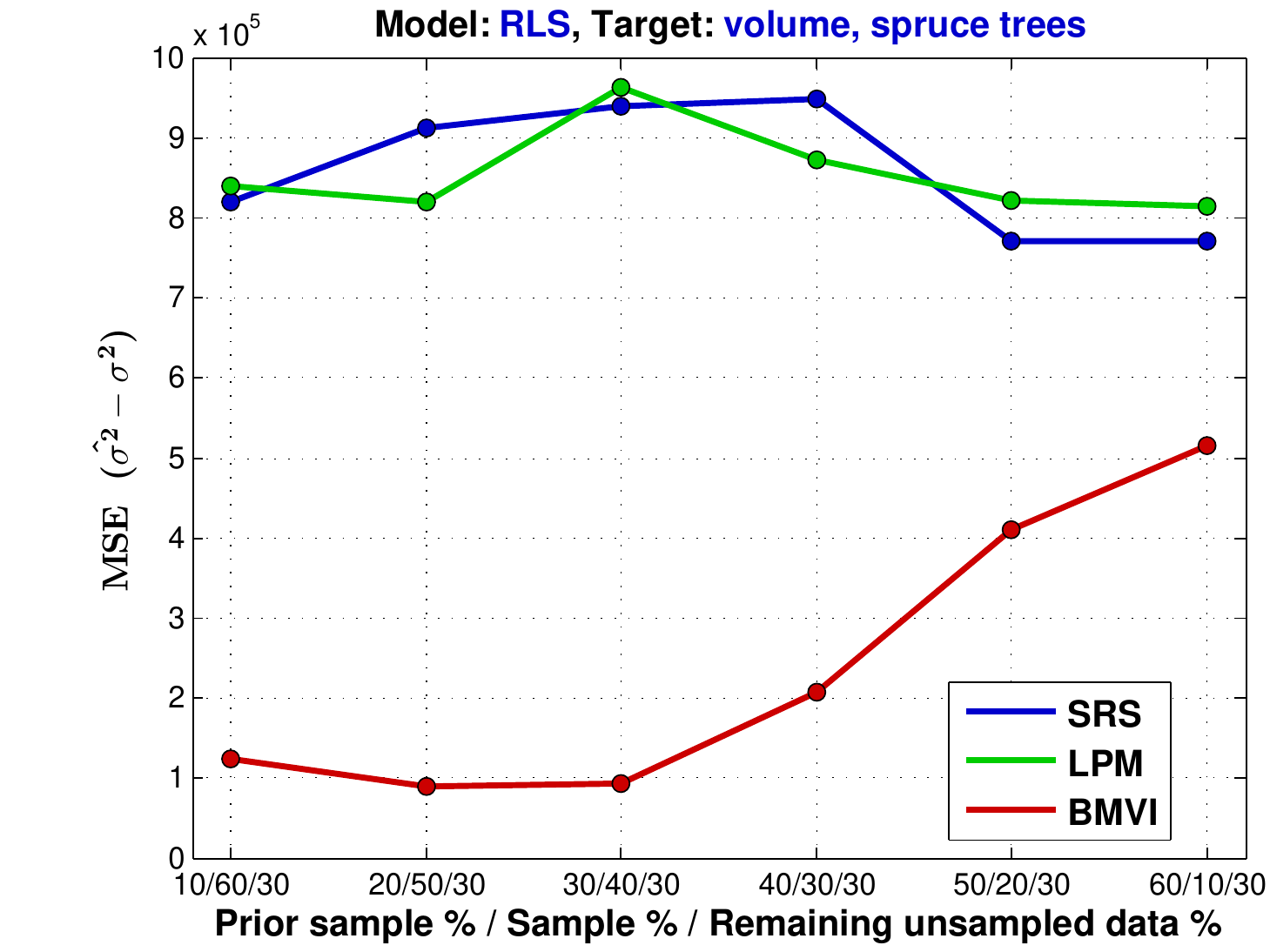}}%

\subfigure[]{\includegraphics[width=.47\textwidth]{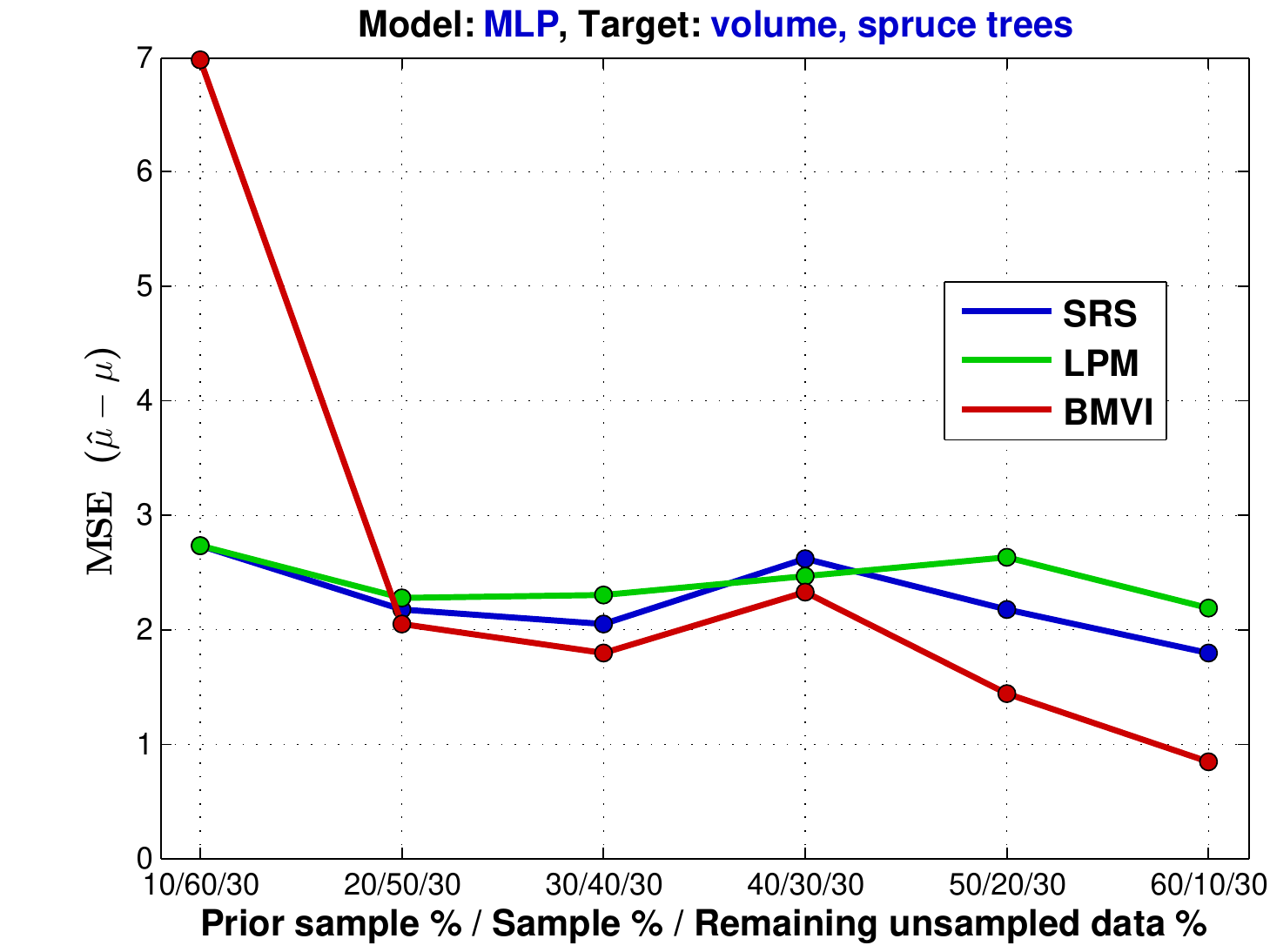}}%
\hspace{1mm}
\subfigure[]{\includegraphics[width=.47\textwidth]{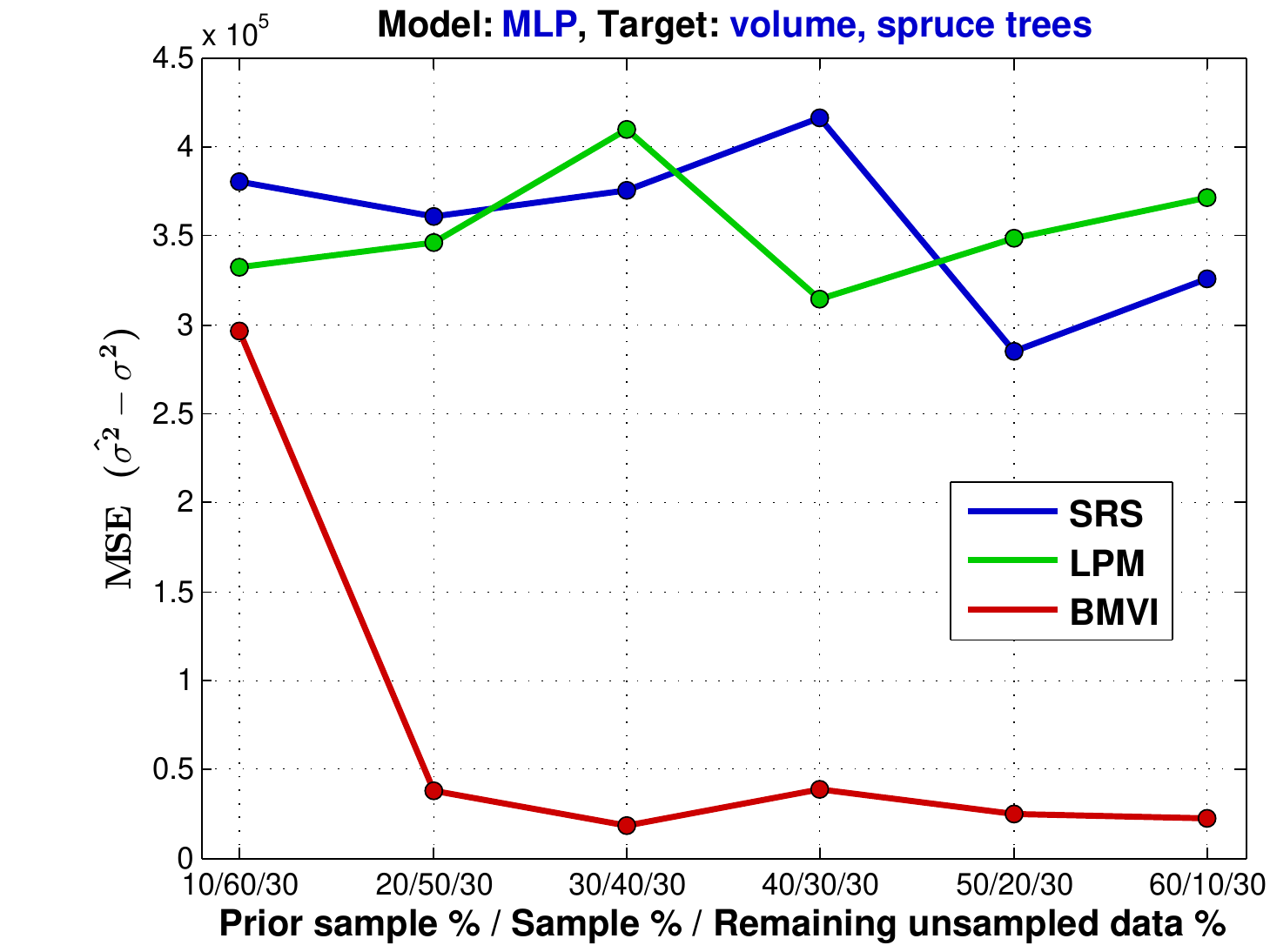}}%

\caption{Results of the empirical comparison of the \BM against SRS and LPM for the response variable: total volume of growing stock, spruce trees. (a)-(b): Population mean and variance estimation performance plots for RLS prediction model. The x-axis represents different values in the fraction vector $\fracset$ and y-axis represents the mean squared error value between estimated and true population parameters. (c)-(d): Analogous results as in (a)-(b) but for MLP prediction model.}
\label{Figure::Analysis_results_volspruce}
\end{figure}

\begin{figure}[b!]
\centering
\subfigure[]{\includegraphics[width=.47\textwidth]{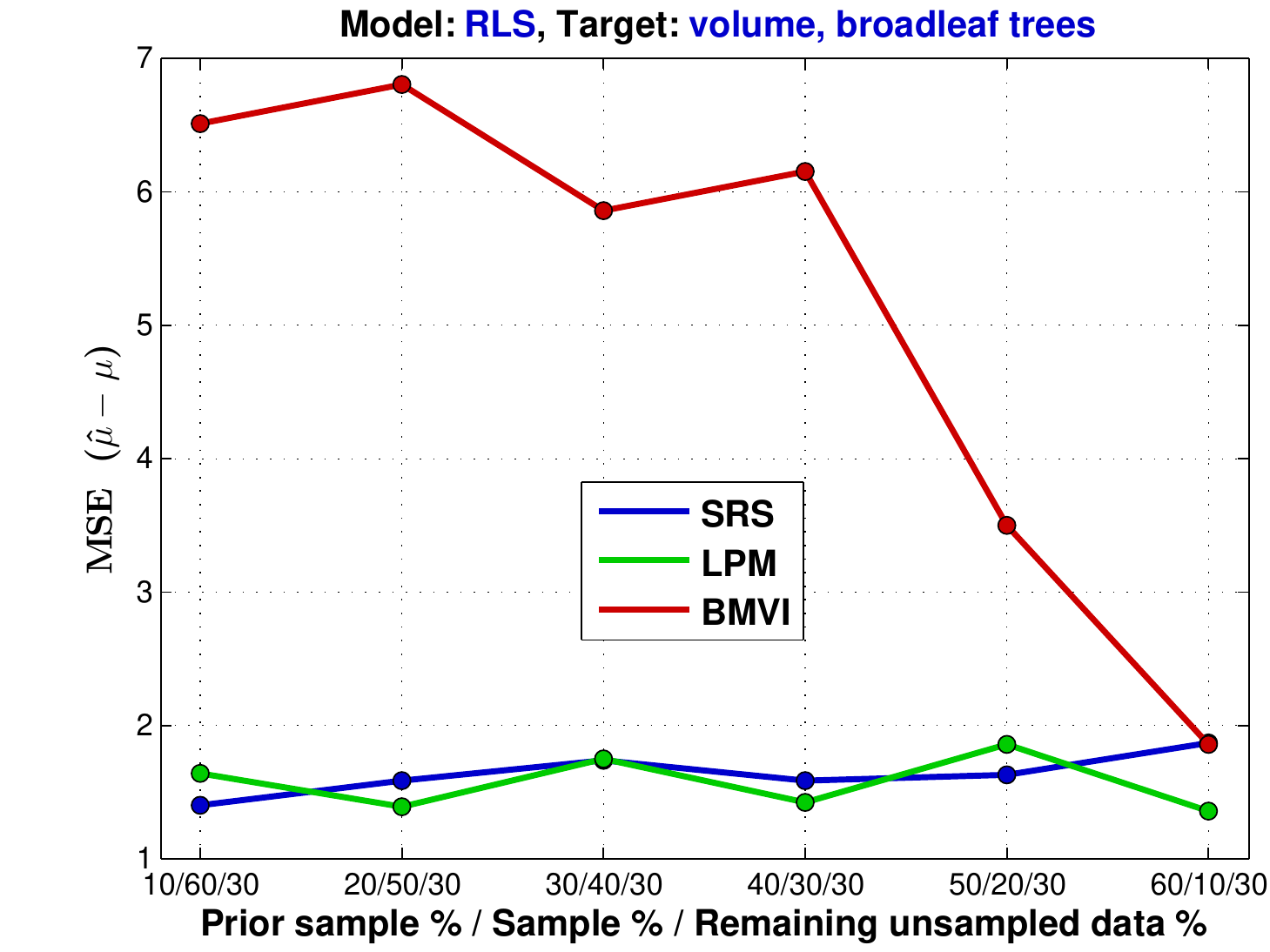}}%
\hspace{1mm}
\subfigure[]{\includegraphics[width=.47\textwidth]{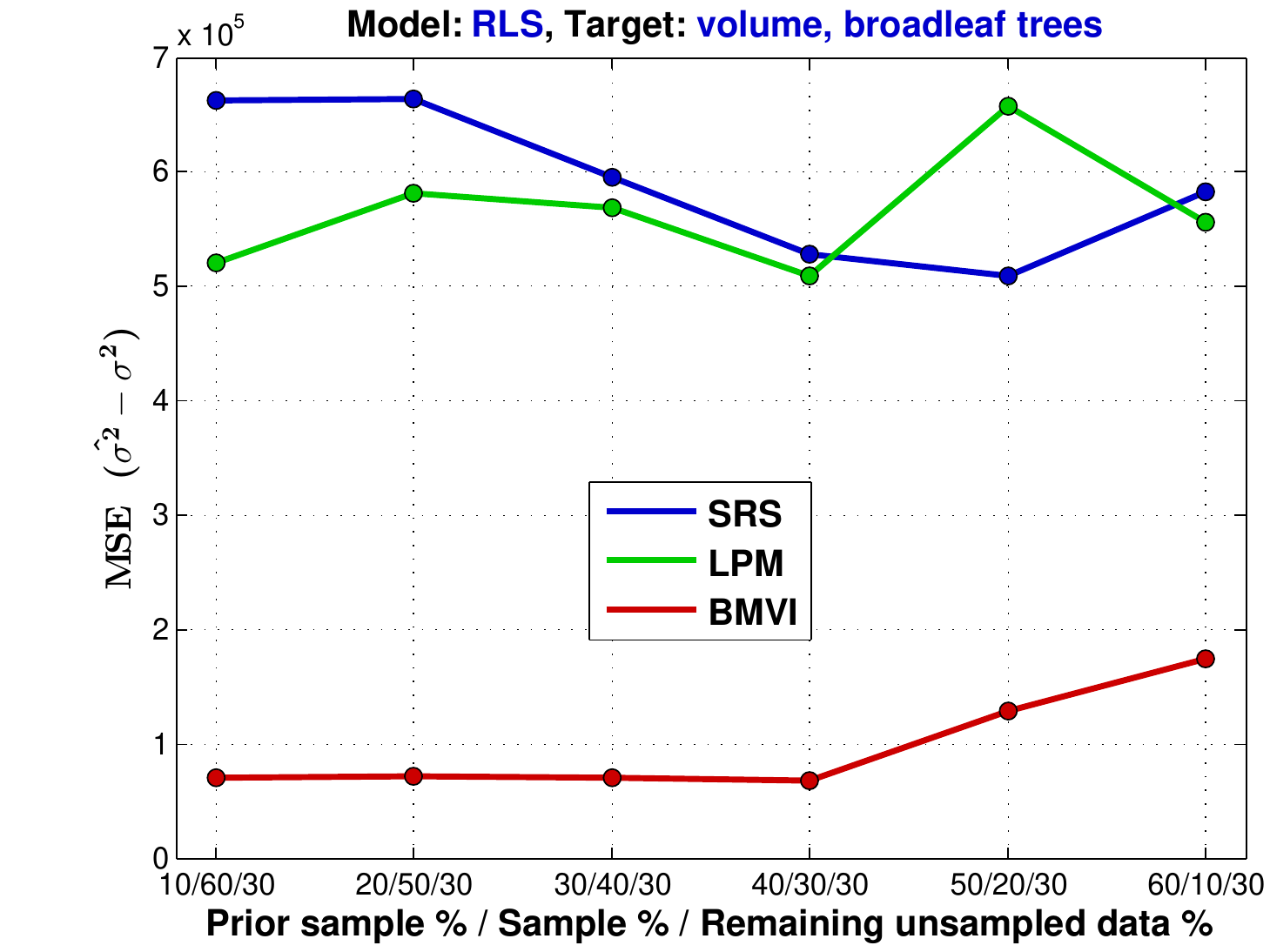}}%

\subfigure[]{\includegraphics[width=.47\textwidth]{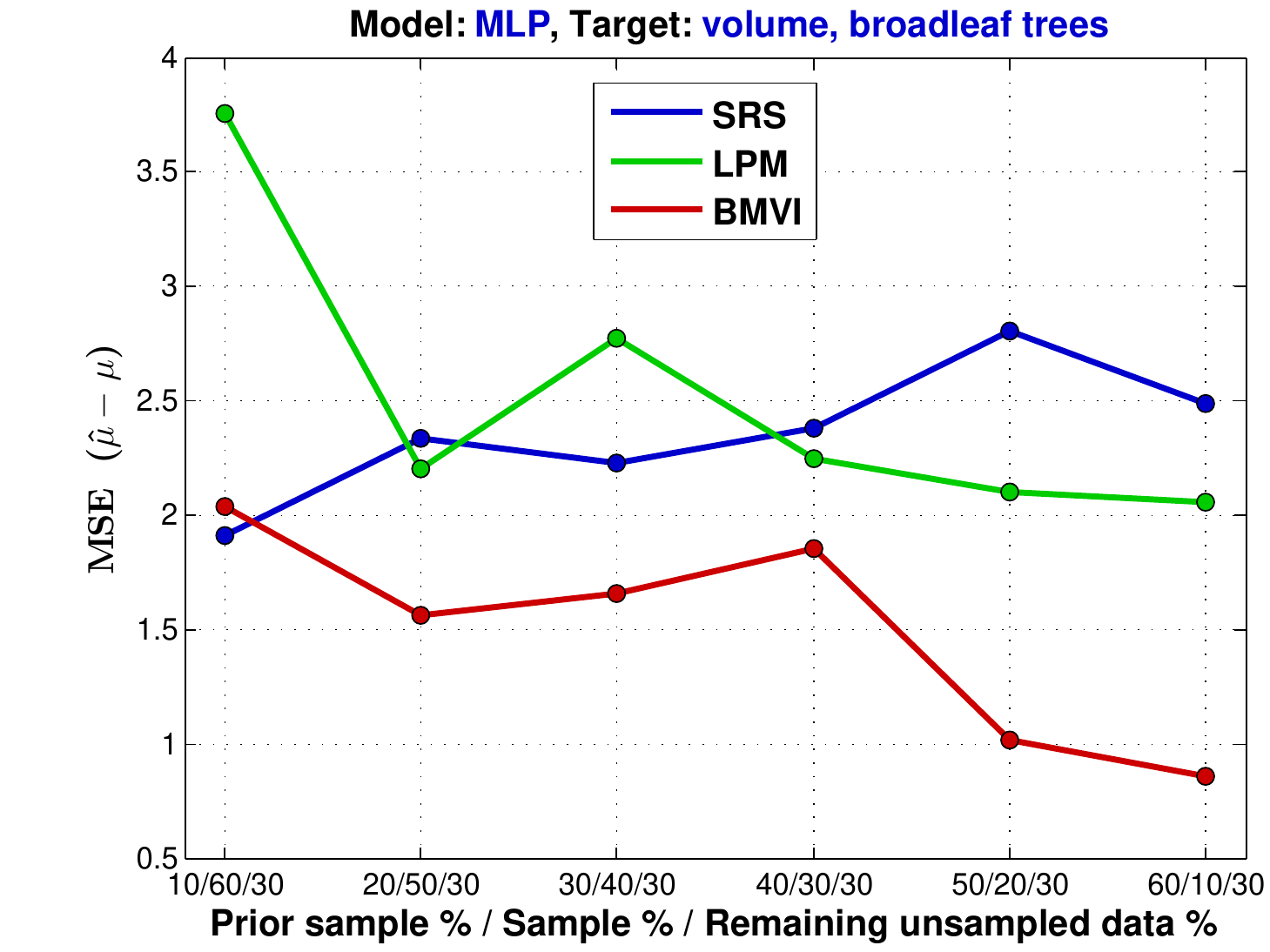}}%
\hspace{1mm}
\subfigure[]{\includegraphics[width=.47\textwidth]{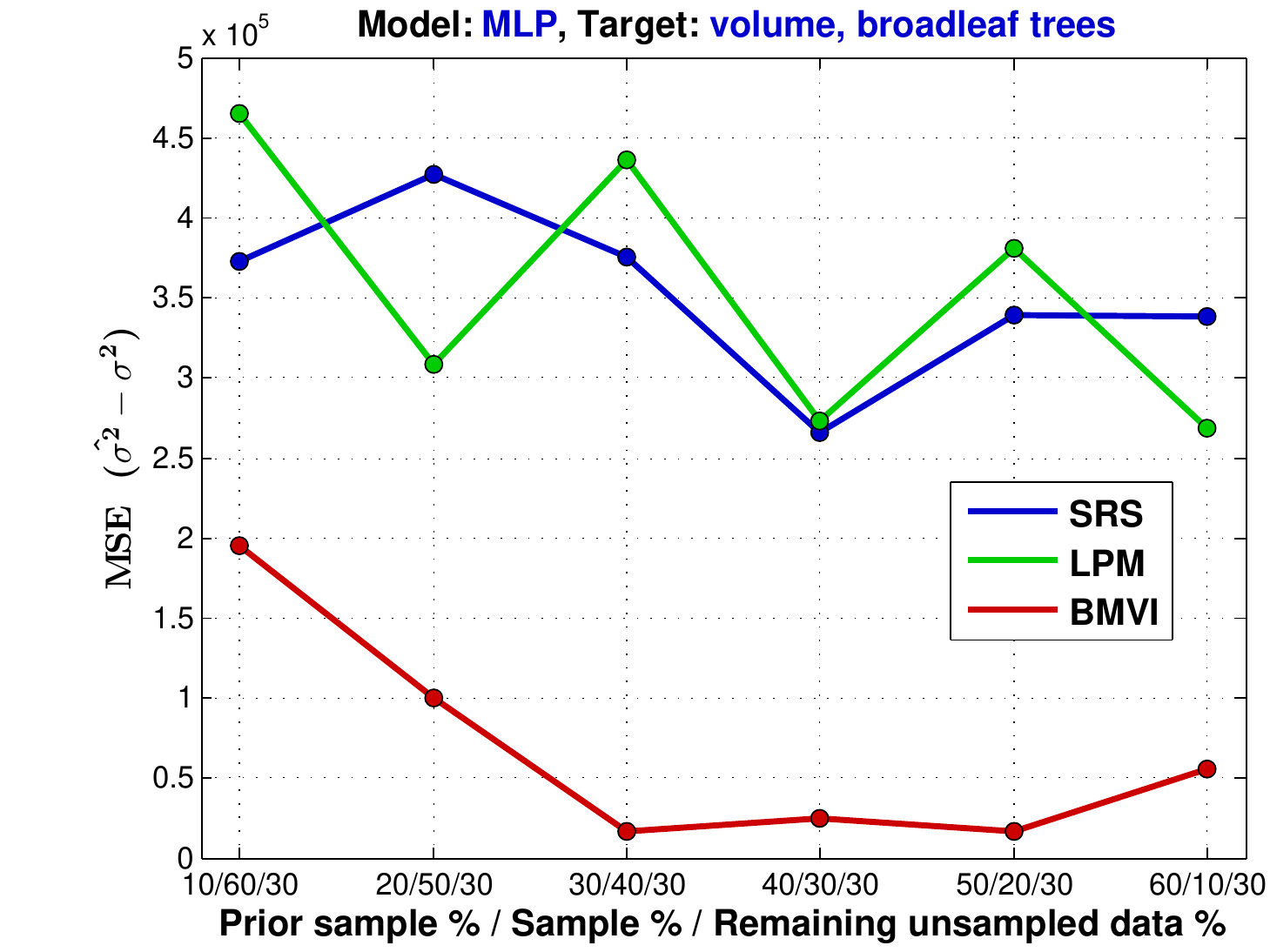}}%

\caption{Results of the empirical comparison of the \BM against SRS and LPM for the response variable: total volume of growing stock, broadleaf trees. (a)-(b): Population mean and variance estimation performance plots for RLS prediction model. The x-axis represents different values in the fraction vector $\fracset$ and y-axis represents the mean squared error value between estimated and true population parameters. (c)-(d): Analogous results as in (a)-(b) but for MLP prediction model.}
\label{Figure::Analysis_results_volbroadleaf}
\end{figure}

\subsection{Case 2: volume of growing stock data}
In Figure \ref{Figure::Analysis_results_volall} we have illustrated the corresponding empirical results for the response variable: volume, all trees (growing stock). As it was with the case of synthetic data, the BMVI model achieves best results with the RLS prediction model for all values of vector $\fracset$. The BMVI method especially well estimates the population variance, both with RLS and MLP prediction models. For the MLP prediction model, the BMVI shows a decreasing MSE value as the number of prior data increases. Thus we can again see the importance of selecting the prediction model correctly. In Figure \ref{Figure::Analysis_results_volpine} we have illustrated the analogous results for the response variable: volume, pine trees (growing stock). Also in this case, the BMVI on average performs best from the three sampling methods when using a RLS prediction model. For a MLP prediction model, the BMVI does not outperform SRS or LPM except in the estimation of population variance. As with the previous response variable case, the BMVI performs better in population variance estimation than with population mean estimation. This is intuitive regarding the design of the BMVI method, which chooses data sample points with high posterior predictive variances. In Figures \ref{Figure::Analysis_results_volspruce} and \ref{Figure::Analysis_results_volbroadleaf} we have illustrated the corresponding results for the response variables: volume, spruce trees and volume, broadleaf trees (growing stock). As a change to the previous results, we now see that in these cases the BMVI method produces best results when using a MLP prediction model. Still in both of these cases, the BMVI again has the best results in estimating population variances.  

We have summarized all the previous results in Tables \ref{Table::results_mean} and \ref{Table::results_variance}. The leftmost columns in the tables represent different valued vectors $\fracset$ (i.e. fraction of data in $\dset_p, \dset_s$ and in the unsampled data set $V$). For example the values .1/.6/.3 in the first row mean that 10\% of the data in population $\mathcal{U}$ is assumed to be known beforehand (as we must have some prior data for the BMVI), 60\% of the data will be sampled from the population, and the remaining 30\% is estimated using the observed data $\dset = \dset_p\cup\dset_s$ and prediction model $f$. We can see the BMVI performing best for response variables synt, $v_a$ and $v_p$ with a linear prediction model RLS in population mean estimation. For response variables $v_s$ and $v_b$, best results are achived by BMVI with MLP prediction model. In the case of population variance estimation, we see the BMVI performing best of the three sampling methods in almost all cases.

As we noted, the results clearly show the gain in utilizing auxiliary information in data sampling strategy. Note that the LPM has very similar performance with the SRS method, both in the synthetic and real world data cases. This result could be explained with the argument that even though the LPM attempts to sample a wider distribution of input features there is no guarantee, generally speaking, that we obtain also a wider distribution in the response variable by sampling data in this manner. One would need to take into account the joint probability distribution of both the inputs $\datax$ and responses $y$, which the LPM does not consider.

\npdecimalsign{4}
\nprounddigits{2}
\begin{table}[]
\centering
\caption{Results of population mean estimations in terms of MSE. The results are illustrated for all response variables, sampling methods and prediction models. The leftmost column of the table represents different valued sampling fraction vector $\fracset$ introduced in section \ref{Section::implementation_details}. In each group of three (SRS, LPM, BMVI) the best sampling method is emphasized with a bolded MSE value.}
\label{Table::results_mean}
  \resizebox{\textwidth}{!}{ 
\begin{tabular}{
>{\columncolor[HTML]{FFFFFF}}c 
>{\columncolor[HTML]{FFFFFF}}c 
>{\columncolor[HTML]{EFEFEF}}c 
>{\columncolor[HTML]{FFFFFF}}c 
>{\columncolor[HTML]{EFEFEF}}c 
>{\columncolor[HTML]{FFFFFF}}c 
>{\columncolor[HTML]{EFEFEF}}c |
>{\columncolor[HTML]{FFFFFF}}c 
>{\columncolor[HTML]{EFEFEF}}c 
>{\columncolor[HTML]{FFFFFF}}c 
>{\columncolor[HTML]{EFEFEF}}c 
>{\columncolor[HTML]{FFFFFF}}c }
 &  & \multicolumn{5}{c|}{\cellcolor[HTML]{FFFFFF}Regularized least squares} & \multicolumn{5}{c}{\cellcolor[HTML]{FFFFFF}Multilayer perceptron} \\ \cline{3-12} 
 & \multicolumn{1}{c|}{\cellcolor[HTML]{FFFFFF}} & \cellcolor[HTML]{FFFFFF}$v_a$ & $v_p$ & \cellcolor[HTML]{FFFFFF}$v_s$ & $v_b$ & \cellcolor[HTML]{FFFFFF}synt & $v_a$ & \cellcolor[HTML]{FFFFFF}$v_p$ & $v_s$ & \cellcolor[HTML]{FFFFFF}$v_b$ & \multicolumn{1}{c|}{\cellcolor[HTML]{FFFFFF}synt} \\ \hline
\multicolumn{1}{c|}{\cellcolor[HTML]{FFFFFF}} & \multicolumn{1}{c|}{\cellcolor[HTML]{FFFFFF}SRS} & 1.51 & 2.829 & 1.794 & \textbf{1.399} & 2.22e-7 & \textbf{1.778} & 3.176 & \textbf{2.732} & \textbf{1.911} & \multicolumn{1}{c|}{\cellcolor[HTML]{FFFFFF}8.69e-5} \\
\multicolumn{1}{c|}{\cellcolor[HTML]{FFFFFF}.1/.6/.3} & \multicolumn{1}{c|}{\cellcolor[HTML]{FFFFFF}LPM} & 1.198 & 2.595 & \textbf{1.382} & 1.639 & \textbf{2.16e-7} & 1.973 & \textbf{2.361} & 2.735 & 3.757 & \multicolumn{1}{c|}{\cellcolor[HTML]{FFFFFF}\textbf{7.17e-5}} \\
\multicolumn{1}{c|}{\cellcolor[HTML]{FFFFFF}} & \multicolumn{1}{c|}{\cellcolor[HTML]{FFFFFF}BMVI} & \textbf{0.832} & \textbf{2.036} & 20.065 & 6.502 & 3.04e-7 & 11.086 & 16.368 & 6.977 & 2.036 & \multicolumn{1}{c|}{\cellcolor[HTML]{FFFFFF}22.52e-5} \\ \hline
\multicolumn{1}{c|}{\cellcolor[HTML]{FFFFFF}} & \multicolumn{1}{c|}{\cellcolor[HTML]{FFFFFF}SRS} & 1.353 & 3.072 & 1.642 & 1.583 & 2.28e-7 & 1.913 & 2.349 & 2.18 & 2.333 & \multicolumn{1}{c|}{\cellcolor[HTML]{FFFFFF}9.14e-5} \\
\multicolumn{1}{c|}{\cellcolor[HTML]{FFFFFF}.2/.5/.3} & \multicolumn{1}{c|}{\cellcolor[HTML]{FFFFFF}LPM} & 1.229 & 2.044 & \textbf{1.307} & \textbf{1.384} & \textbf{2.21e-7} & \textbf{1.585} & \textbf{2.181} & 2.276 & 2.202 & \multicolumn{1}{c|}{\cellcolor[HTML]{FFFFFF}\textbf{8.57e-5}} \\
\multicolumn{1}{c|}{\cellcolor[HTML]{FFFFFF}} & \multicolumn{1}{c|}{\cellcolor[HTML]{FFFFFF}BMVI} & \textbf{0.796} & \textbf{1.43} & 19.881 & 6.796 & 2.81e-7 & 7.03 & 16.576 & \textbf{2.054} & \textbf{1.561} & \multicolumn{1}{c|}{\cellcolor[HTML]{FFFFFF}15.5e-5} \\ \hline
\multicolumn{1}{c|}{\cellcolor[HTML]{FFFFFF}} & \multicolumn{1}{c|}{\cellcolor[HTML]{FFFFFF}SRS} & 1.536 & 2.502 & 2.111 & \textbf{1.731} & 2.36e-7 & \textbf{2.323} & \textbf{2.416} & 2.047 & 2.226 & \multicolumn{1}{c|}{\cellcolor[HTML]{FFFFFF}9.59e-5} \\
\multicolumn{1}{c|}{\cellcolor[HTML]{FFFFFF}.3/.4/.3} & \multicolumn{1}{c|}{\cellcolor[HTML]{FFFFFF}LPM} & 1.733 & 2.589 & \textbf{1.851} & 1.747 & \textbf{2.25e-7} & 2.61 & 2.439 & 2.303 & 2.774 & \multicolumn{1}{c|}{\cellcolor[HTML]{FFFFFF}\textbf{7.81e-5}} \\
\multicolumn{1}{c|}{\cellcolor[HTML]{FFFFFF}} & \multicolumn{1}{c|}{\cellcolor[HTML]{FFFFFF}BMVI} & \textbf{1.046} & \textbf{2.158} & 17.148 & 5.848 & 2.33e-7 & 11.222 & 9.252 & \textbf{1.8} & \textbf{1.659} & \multicolumn{1}{c|}{\cellcolor[HTML]{FFFFFF}12.18e-5} \\ \hline
\multicolumn{1}{c|}{\cellcolor[HTML]{FFFFFF}} & \multicolumn{1}{c|}{\cellcolor[HTML]{FFFFFF}SRS} & 1.381 & 2.083 & 2.085 & 1.579 & 2.28e-7 & \textbf{1.935} & 2.077 & 2.612 & 2.378 & \multicolumn{1}{c|}{\cellcolor[HTML]{FFFFFF}9.21e-5} \\
\multicolumn{1}{c|}{\cellcolor[HTML]{FFFFFF}.4/.3/.3} & \multicolumn{1}{c|}{\cellcolor[HTML]{FFFFFF}LPM} & 1.241 & 2.416 & \textbf{1.648} & \textbf{1.414} & 2.31e-7 & 2.006 & \textbf{1.852} & 2.472 & 2.248 & \multicolumn{1}{c|}{\cellcolor[HTML]{FFFFFF}\textbf{7.68e-5}} \\
\multicolumn{1}{c|}{\cellcolor[HTML]{FFFFFF}} & \multicolumn{1}{c|}{\cellcolor[HTML]{FFFFFF}BMVI} & \textbf{0.945} & \textbf{2.037} & 12.111 & 6.148 & \textbf{1.79e-7} & 3.024 & 5.594 & \textbf{2.332} & \textbf{1.853} & \multicolumn{1}{c|}{\cellcolor[HTML]{FFFFFF}12.78e-5} \\ \hline
\multicolumn{1}{c|}{\cellcolor[HTML]{FFFFFF}} & \multicolumn{1}{c|}{\cellcolor[HTML]{FFFFFF}SRS} & 1.588 & 2.434 & 1.812 & \textbf{1.623} & 2.34e-7 & 1.851 & 2.638 & 2.176 & 2.804 & \multicolumn{1}{c|}{\cellcolor[HTML]{FFFFFF}9.02e-5} \\
\multicolumn{1}{c|}{\cellcolor[HTML]{FFFFFF}.5/.2/.3} & \multicolumn{1}{c|}{\cellcolor[HTML]{FFFFFF}LPM} & 1.184 & \textbf{2.195} & \textbf{1.569} & 1.853 & 2.33e-7 & \textbf{1.472} & \textbf{1.955} & 2.638 & 2.104 & \multicolumn{1}{c|}{\cellcolor[HTML]{FFFFFF}\textbf{8.77e-5}} \\
\multicolumn{1}{c|}{\cellcolor[HTML]{FFFFFF}} & \multicolumn{1}{c|}{\cellcolor[HTML]{FFFFFF}BMVI} & \textbf{0.871} & 2.321 & 6.851 & 3.495 & \textbf{1.38e-7} & 2.621 & 3.756 & \textbf{1.443} & \textbf{1.015} & \multicolumn{1}{c|}{\cellcolor[HTML]{FFFFFF}10.78e-5} \\ \hline
\multicolumn{1}{c|}{\cellcolor[HTML]{FFFFFF}} & \multicolumn{1}{c|}{\cellcolor[HTML]{FFFFFF}SRS} & 1.211 & 3.615 & \textbf{1.613} & 1.861 & 2.3e-7 & \textbf{1.824} & 2.811 & 1.795 & 2.491 & \multicolumn{1}{c|}{\cellcolor[HTML]{FFFFFF}9.33e-5} \\
\multicolumn{1}{c|}{\cellcolor[HTML]{FFFFFF}.6/.1/.3} & \multicolumn{1}{c|}{\cellcolor[HTML]{FFFFFF}LPM} & 1.184 & 2.996 & 1.83 & \textbf{1.353} & 2.31e-7 & 2.494 & \textbf{2.341} & 2.186 & 2.059 & \multicolumn{1}{c|}{\cellcolor[HTML]{FFFFFF}\textbf{7.26e-5}} \\
\multicolumn{1}{c|}{\cellcolor[HTML]{FFFFFF}} & \multicolumn{1}{c|}{\cellcolor[HTML]{FFFFFF}BMVI} & \textbf{1.094} & \textbf{1.641} & 2.761 & 1.855 & \textbf{1.84e-7} & 2.782 & 4.221 & \textbf{0.844} & \textbf{0.861} & \multicolumn{1}{c|}{\cellcolor[HTML]{FFFFFF}8.92e-5} \\ \hline
\end{tabular}
}
\end{table}
\npnoround
\npdecimalsign{4}
\nprounddigits{2}
\begin{table}[t!]
\centering
\caption{Analogous results as in Table \ref{Table::results_mean} but for population variance estimations.}
\label{Table::results_variance}
  \resizebox{\textwidth}{!}{ 
\begin{tabular}{
>{\columncolor[HTML]{FFFFFF}}c 
>{\columncolor[HTML]{FFFFFF}}c 
>{\columncolor[HTML]{EFEFEF}}c 
>{\columncolor[HTML]{FFFFFF}}c 
>{\columncolor[HTML]{EFEFEF}}c 
>{\columncolor[HTML]{FFFFFF}}c 
>{\columncolor[HTML]{EFEFEF}}c |
>{\columncolor[HTML]{FFFFFF}}c 
>{\columncolor[HTML]{EFEFEF}}c 
>{\columncolor[HTML]{FFFFFF}}c 
>{\columncolor[HTML]{EFEFEF}}c 
>{\columncolor[HTML]{FFFFFF}}c }
 &  & \multicolumn{5}{c|}{\cellcolor[HTML]{FFFFFF}Regularized least squares} & \multicolumn{5}{c}{\cellcolor[HTML]{FFFFFF}Multilayer perceptron} \\ \cline{3-12} 
 & \multicolumn{1}{c|}{\cellcolor[HTML]{FFFFFF}} & \cellcolor[HTML]{FFFFFF}$v_a$ & $v_p$ & \cellcolor[HTML]{FFFFFF}$v_s$ & $v_b$ & \cellcolor[HTML]{FFFFFF}synt & $v_a$ & \cellcolor[HTML]{FFFFFF}$v_p$ & $v_s$ & \cellcolor[HTML]{FFFFFF}$v_b$ & \multicolumn{1}{c|}{\cellcolor[HTML]{FFFFFF}synt} \\ \hline
\multicolumn{1}{c|}{\cellcolor[HTML]{FFFFFF}} & \multicolumn{1}{c|}{\cellcolor[HTML]{FFFFFF}SRS} & 6.25e+5 & 8.67e+5 & 8.2e+5 & 6.63e+5 & 1.3e-11 & \textbf{7.55e+5} & \textbf{4.97e+5} & 3.8e+5 & 3.73e+5 & \multicolumn{1}{c|}{\cellcolor[HTML]{FFFFFF}7.11e-5} \\
\multicolumn{1}{c|}{\cellcolor[HTML]{FFFFFF}.1/.6/.3} & \multicolumn{1}{c|}{\cellcolor[HTML]{FFFFFF}LPM} & 7.83e+5 & 9.07e+5 & 8.4e+5 & 5.2e+5 & 1.27e-11 & 9.88e+5 & 6.82e+5 & 3.32e+5 & 4.65e+5 & \multicolumn{1}{c|}{\cellcolor[HTML]{FFFFFF}\textbf{5.9e-5}} \\
\multicolumn{1}{c|}{\cellcolor[HTML]{FFFFFF}} & \multicolumn{1}{c|}{\cellcolor[HTML]{FFFFFF}BMVI} & \textbf{0.24e+5} & \textbf{5.5e+5} & \textbf{1.24e+5} & \textbf{0.7e+5} & \textbf{0.55e-11} & 12.2e+5 & 9.34e+5 & \textbf{2.96e+5} & \textbf{1.95e+5} & \multicolumn{1}{c|}{\cellcolor[HTML]{FFFFFF}18.5e-5} \\ \hline
\multicolumn{1}{c|}{\cellcolor[HTML]{FFFFFF}} & \multicolumn{1}{c|}{\cellcolor[HTML]{FFFFFF}SRS} & 7.29e+5 & 8.66e+5 & 9.12e+5 & 6.64e+5 & 1.4e-11 & 9.56e+5 & 5.91e+5 & 3.61e+5 & 4.27e+5 & \multicolumn{1}{c|}{\cellcolor[HTML]{FFFFFF}7.69e-5} \\
\multicolumn{1}{c|}{\cellcolor[HTML]{FFFFFF}.2/.5/.3} & \multicolumn{1}{c|}{\cellcolor[HTML]{FFFFFF}LPM} & 6.98e+5 & 9.38e+5 & 8.2e+5 & 5.82e+5 & 1.31e-11 & 9.95e+5 & \textbf{5.44e+5} & 3.46e+5 & 3.08e+5 & \multicolumn{1}{c|}{\cellcolor[HTML]{FFFFFF}\textbf{7.08e-5}} \\
\multicolumn{1}{c|}{\cellcolor[HTML]{FFFFFF}} & \multicolumn{1}{c|}{\cellcolor[HTML]{FFFFFF}BMVI} & \textbf{0.18e+5} & \textbf{5.18e+5} & \textbf{0.89e+5} & \textbf{0.71e+5} & \textbf{0.54e-11} & \textbf{6.9e+5} & 7.96e+5 & \textbf{0.38e+5} & \textbf{0.99e+5} & \multicolumn{1}{c|}{\cellcolor[HTML]{FFFFFF}12.8e-5} \\ \hline
\multicolumn{1}{c|}{\cellcolor[HTML]{FFFFFF}} & \multicolumn{1}{c|}{\cellcolor[HTML]{FFFFFF}SRS} & 5.7e+5 & 8.38e+5 & 9.4e+5 & 5.95e+5 & 1.37e-11 & 8.27e+5 & 5.23e+5 & 3.75e+5 & 3.76e+5 & \multicolumn{1}{c|}{\cellcolor[HTML]{FFFFFF}7.97e-5} \\
\multicolumn{1}{c|}{\cellcolor[HTML]{FFFFFF}.3/.4/.3} & \multicolumn{1}{c|}{\cellcolor[HTML]{FFFFFF}LPM} & 8.5e+5 & 8.35e+5 & 9.63e+5 & 5.68e+5 & 1.33e-11 & 9.74e+5 & 5.79e+5 & 4.1e+5 & 4.36e+5 & \multicolumn{1}{c|}{\cellcolor[HTML]{FFFFFF}\textbf{6.42e-5}} \\
\multicolumn{1}{c|}{\cellcolor[HTML]{FFFFFF}} & \multicolumn{1}{c|}{\cellcolor[HTML]{FFFFFF}BMVI} & \textbf{0.26e+5} & \textbf{5.51e+5} & \textbf{0.94e+5} & \textbf{0.7e+5} & \textbf{0.56e-11} & \textbf{4.24e+5} & \textbf{3.5e+5} & \textbf{0.18e+5} & \textbf{0.17e+5} & \multicolumn{1}{c|}{\cellcolor[HTML]{FFFFFF}10e-5} \\ \hline
\multicolumn{1}{c|}{\cellcolor[HTML]{FFFFFF}} & \multicolumn{1}{c|}{\cellcolor[HTML]{FFFFFF}SRS} & 4.36e+5 & 9.24e+5 & 9.49e+5 & 5.28e+5 & 1.43e-11 & 7.49e+5 & 5.9e+5 & 4.17e+5 & 2.66e+5 & \multicolumn{1}{c|}{\cellcolor[HTML]{FFFFFF}7.72e-5} \\
\multicolumn{1}{c|}{\cellcolor[HTML]{FFFFFF}.4/.3/.3} & \multicolumn{1}{c|}{\cellcolor[HTML]{FFFFFF}LPM} & 5.73e+5 & 7.92e+5 & 8.72e+5 & 5.09e+5 & 1.35e-11 & 6.75e+5 & 5.52e+5 & 3.15e+5 & 2.73e+5 & \multicolumn{1}{c|}{\cellcolor[HTML]{FFFFFF}\textbf{6.22e-5}} \\
\multicolumn{1}{c|}{\cellcolor[HTML]{FFFFFF}} & \multicolumn{1}{c|}{\cellcolor[HTML]{FFFFFF}BMVI} & \textbf{0.44e+5} & \textbf{5.26e+5} & \textbf{2.08e+5} & \textbf{0.68e+5} & \textbf{0.69e-11} & \textbf{3.94e+5} & \textbf{3.6e+5} & \textbf{0.39e+5} & \textbf{0.25e+5} & \multicolumn{1}{c|}{\cellcolor[HTML]{FFFFFF}10.6e-5} \\ \hline
\multicolumn{1}{c|}{\cellcolor[HTML]{FFFFFF}} & \multicolumn{1}{c|}{\cellcolor[HTML]{FFFFFF}SRS} & 5.53e+5 & 9.38e+5 & 7.71e+5 & 5.09e+5 & 1.41e-11 & 7.09e+5 & 5.79e+5 & 2.85e+5 & 3.39e+5 & \multicolumn{1}{c|}{\cellcolor[HTML]{FFFFFF}7.67e-5} \\
\multicolumn{1}{c|}{\cellcolor[HTML]{FFFFFF}.5/.2/.3} & \multicolumn{1}{c|}{\cellcolor[HTML]{FFFFFF}LPM} & 5.28e+5 & 8.42e+5 & 8.21e+5 & 6.58e+5 & 1.43e-11 & 8.53e+5 & 5.61e+5 & 3.49e+5 & 3.81e+5 & \multicolumn{1}{c|}{\cellcolor[HTML]{FFFFFF}\textbf{7.38e-5}} \\
\multicolumn{1}{c|}{\cellcolor[HTML]{FFFFFF}} & \multicolumn{1}{c|}{\cellcolor[HTML]{FFFFFF}BMVI} & \textbf{0.37e+5} & \textbf{5.42e+5} & \textbf{4.1e+5} & \textbf{1.29e+5} & \textbf{0.96e-11} & \textbf{2.36e+5} & \textbf{2.49e+5} & \textbf{0.25e+5} & \textbf{0.16e+5} & \multicolumn{1}{c|}{\cellcolor[HTML]{FFFFFF}9.05e-5} \\ \hline
\multicolumn{1}{c|}{\cellcolor[HTML]{FFFFFF}} & \multicolumn{1}{c|}{\cellcolor[HTML]{FFFFFF}SRS} & 5.26e+5 & 8.45e+5 & 7.71e+5 & 5.83e+5 & 1.37e-11 & 10e+5 & 5.1e+5 & 3.25e+5 & 3.38e+5 & \multicolumn{1}{c|}{\cellcolor[HTML]{FFFFFF}8.01e-5} \\
\multicolumn{1}{c|}{\cellcolor[HTML]{FFFFFF}.6/.1/.3} & \multicolumn{1}{c|}{\cellcolor[HTML]{FFFFFF}LPM} & 4.4e+5 & 9.17e+5 & 8.14e+5 & 5.56e+5 & 1.37e-11 & 12e+5 & 6.34e+5 & 3.71e+5 & 2.68e+5 & \multicolumn{1}{c|}{\cellcolor[HTML]{FFFFFF}\textbf{6.11e-5}} \\
\multicolumn{1}{c|}{\cellcolor[HTML]{FFFFFF}} & \multicolumn{1}{c|}{\cellcolor[HTML]{FFFFFF}BMVI} & \textbf{0.47e+5} & \textbf{6.02e+5} & \textbf{5.15e+5} & \textbf{1.74e+5} & \textbf{1.1e-11} & \textbf{1.7e+5} & \textbf{2.95e+5} & \textbf{0.22e+5} & \textbf{0.55e+5} & \multicolumn{1}{c|}{\cellcolor[HTML]{FFFFFF}7.38e-5} \\ \hline
\end{tabular}
}
\end{table}
\npnoround

\pagebreak

\section{Discussion} \label{Section::Discussion}
Regarding the experiments conducted in this study, similar researches utilizing auxiliary information in forest inventories have been done done e.g. in the works of \citep{Raty2018,Grafstrom2013}. In the work of \citep{Raty2018} the authors showed significant improvements in sampling efficiency for forest inventories with the usage of auxiliary remote sensing data. Also, in \citep{Grafstrom2013} experiments made with both synthetic and real data showed great utility of using airborne laser scanning data in forest inventory sampling design. Furthermore, the application of Bayesian approaches in optimizing geostatistical sampling designs can be found from a variety of literature, e.g. in the works by \citep{Werner2012,Muller2007,Zhu2006,Diggle2006}.

The empirical results of this study showed best performance for the BMVI method when a RLS prediction model was used in the cases of synthetic, volume all trees, and volume pine trees response data. Correspondingly, the best results were achieved by the BMVI method in the cases of volume spruce and broadleaf response data when a MLP prediction model was used. The results for SRS and LPM were similar in all data cases. 

The results suggest that one should consider the used prediction model family before applying the \BM sampling method. This is a relevant consideration since the \BM method's data sample inclusion criteria depends on the functional dependency defined by the chosen prediction model $f(\datax; \w)$. Thus, if one forces a specific prior model family between the auxiliary predictor and response data sets without justification, then one risks on utilizing potentially non-existing relationship in the data in sampling decisions. For example, the empirical results in figures \ref{Figure::Analysis_results_volspruce} and \ref{Figure::Analysis_results_volbroadleaf} suggest a non-RLS relationship between the predictor data and spruce/broadleaf tree. To summarize the point, one should investigate and justify the selection of a particular prediction model family before applying the \BM method in order to obtain optimal performance. The results with synthetic data set also verified the best performance for the \BM method when the relevant assumptions were satisfied.

In this work, the sampling methods were compared by measuring their capability in producing estimates for the response data population parameters, namely the mean and variance. It is worth mentioning, that in a completely general forest inventory case the BMVI probably does not outperform SRS or LPM in population parameter estimation. This is can be obviously caused by the unavailability of sufficient prior data or the lacking of a learnable functional relationship between the auxiliary and response data, on which the BMVI relies on. However, if the previous requirements are satisfied then the results revealed that improved sampling decisions can be obtained with the BMVI when compared with SRS and LPM. The main utility of the BMVI is in utilizing the information gained from already sampled data in new sample designs. 

While forest inventories such as NFIs usually contain information on hundreds of variables, the univariate results presented in this work were focused on tree volume variables, since the sampling design is optimized for this purpose in Finland. We can also see this in the studies by \citep{Raty2018,Grafstrom2013} where tree volume has been the main variable of interest. The corresponding sampling design optimized for tree volume will also be used also for all other variables recorded in the NFI, since it would be practically infeasible to optimize the sampling design for all variables of interest. It is up to the user of the BMVI method to decide which forest inventory variables to give main weight in the sampling design. 

Lastly, we wish to state the limiting factor of the real world data used in this study. For an improved analysis, the simulated sampling results of this study should be repeated with a real world data set with all the response data known throughout the research area. In this work, the real world data was available only as a systematic cluster sample from the research area. However, the promising experimental results encourage the continued future analysis of the BMVI method in the context of remote sensing-based forest inventories, which is also supported by the results produced with the synthetic data.

\section{Conclusions}\label{Section::conclusions}
In this study, we proposed the data sampling method BMVI, which utilizes the information gained from learned relationships between the auxiliary RS and the forest inventory variables. The results revealed that when enough prior data was available together with a suitable modeling approach, best performance in the estimation of response variable population parameters was achieved with the BMVI method, both in the synthetic and real world RS/inventory data set cases. It was thus confirmed that the utilization of RS data in the forest inventory sample selection via Bayesian optimization can improve the population parameter estimation.

\bibliographystyle{elsarticle-harv}
\bibliography{references.bib}

\pagebreak

\section*{Appendix}

In this appendix, we will derive the result in equation \ref{Equation::post_pred_variance_result_beg} of section \ref{Section::BMVI}. We will next proceed with formulating a closed-form expression for $\yvar$. To begin, we assume a Gaussian prior distribution for the prediction model parameters:
\begin{equation}
\label{Equation::w_prior}
p(\w)\propto \exp\left(-\frac{\alpha}{2}||\w||^2\right)=\exp\left(-\frac{\alpha}{2}\sum_{j=1}^q \theta_j^2\right)=\prod_{j=1}^q \exp{\left(-\frac{\alpha}{2}\theta_j^2\right)}.   
\end{equation}
That is, each model parameter $\theta_j$ is assumed to be distributed as $\theta_j \sim \mathcal{N}(0, \alpha^{-1})$. The response variable $y$ is assumed to be generated by a function $f(\datax; \w)$ with additive zero-mean Gaussian noise $\epsilon \sim \mathcal{N}(0, \beta^{-1})$, i.e.: 

\begin{equation}
    p(y|\datax, \w)\propto \exp{\left(-\frac{\beta}{2} \epsilon^2\right)},
    \label{Equation::y_prior_distribution}
\end{equation}
where $\epsilon=y-f(\datax; \w)$. By also assuming that the data set $\dset$ consists from identically and independently distributed samples, we get the data likelihood as: 
\begin{equation}
\label{Equation::data_likelihood}
p(\dset|\w)\propto \prod_{i=1}^N p(y|\datax_i, \w)=\exp{\left(-\frac{\beta}{2}\sum_{i=1}^N \{y_i-f(\datax_i; \w)\}^2\right)}.
\end{equation}
We can now use equations \ref{Equation::w_prior} and \ref{Equation::data_likelihood} to express the posterior distribution for $\w$ as: 
\begin{equation}
\label{Equation::w_posterior}
p(\w|\dset)\propto\, p(\dset|\w)\,p(\w) \propto \exp{\left(-\frac{\beta}{2}\sum_{i=1}^N \{y_i-f(\datax_i; \w)\}^2-\frac{\alpha}{2}\sum_{j=1}^q \theta_j^2\right)}.
\end{equation}
Next, we will denote the negative of the exponent in equation \ref{Equation::w_posterior} as
\begin{equation}
\label{Equation::error_function_S(w)}
S(\w)=\frac{\beta}{2}\sum_{i=1}^N \{y_i-f(\datax_i; \w)\}^2+\frac{\alpha}{2}\sum_{j=1}^q \theta_j^2,  
\end{equation}
and make a second degree Taylor approximation for this function around the maximum posterior point $\wmp=\argmax_{\w\in \mathbb{R}^q} p(\w|\dset)$:
\begin{equation}
S(\w)\approx S(\wmp)+\frac12(\w-\wmp)^T\textbf{A}(\w-\wmp),
\end{equation}
where $\textbf{A}$ is the Hessian matrix of $S(\w)$ evaluated at $\wmp$, i.e. the $(i,j)^{\text{th}}$ element of $\textbf{A}$ is $$\textbf{A}_{i,j}=\frac{\partial^2}{\partial \theta_i \partial \theta_j}\left(S(\w)\right)\rvert_{\w=\wmp}.$$
The maximum posterior $\wmp$ corresponds also to the parameters, which minimize $S(\w)$, i.e. $\wmp = \argmin_{\w\in \mathbb{R}^q} S(\w)$. Furthermore, note that the prior distribution $p(\w)$ provides a regularizing function into $S(\w)$ which results in favoring smaller values of $\theta_j$, thus encouraging the selection of smoother functions $f(\datax; \w)$ in $\wmp$ solution. Assuming in addition that the width of the posterior distribution of $\w$ is sufficiently narrow (due to the Hessian $\textbf{A}$), we can approximate $f(\datax; \w)$ with a linear expansion around $\wmp$ as $f(\datax; \w)\approx f(\datax; \wmp)+\g^T(\w-\wmp)$, where $\g$ is the gradient vector of $f(\datax; \w)$ with respect to $\w$ evaluated at $\wmp$. That is, the $j^{\text{th}}$ element of $\g$ is:
$$g_j= \frac{\partial}{\partial \theta_j}\left(f(\datax; \w)\right)\rvert_{\w=\wmp}.$$
The linear approximation of $f(\datax; \w)$ is suitable here without significantly losing accuracy, since most of the probability mass is focused on $\wmp$ and the higher order terms of the expansion are close to zero. By now plugging equations \ref{Equation::y_prior_distribution} and \ref{Equation::w_posterior} into equation \ref{Equation::posterior_predictive_y}, using the approximations of $S(\w)$ and $f(\datax; \w)$ and denoting $\Delta\w = \w-\wmp$ and $\ymean = f(\datax; \wmp)$, we get the expression for the posterior predictive distribution for $y$ in equation \ref{Equation::posterior_predictive_y} as: 
\begin{equation}
\label{Equation::post_pred_y_ready}
\begin{aligned}
p(y|\datax, \dset) &\propto \int_{\mathbb{R}^q} \exp{\left(-\frac{\beta}{2}\{y-\ymean-\g^T\Delta\w\}^2\right)}\exp{\left(-S(\wmp)-\frac12\Delta\w^T\textbf{A}\Delta\w\right)} \,d\w\\ &\propto \int_{\mathbb{R}^q}\exp{\left(-\frac{\beta}{2}\{y-\ymean-\g^T\Delta\w\}^2-\frac{1}{2}\Delta\w^T\textbf{A}\Delta\w\right)}\,d\w\\
&=(2\pi)^{q/2}\lvert\textbf{A}+\beta\g\g^T\rvert^{-1/2}\exp{\left(-\frac{\{y-\ymean\}^2}{2\sigma^2_{\dset}(\datax)}\right)}, 
\end{aligned}
\end{equation}
where now the variance of posterior predictive distribution of $y$ is:
\begin{equation}
\label{Equation::post_pred_variance_result}
\sigma_{\dset}^2(\datax) = \frac{1}{\beta-\beta^2\g^T \left(\textbf{A}+\beta\g\g^T\right)^{-1}\g} = \frac{1}{\beta} + \g(\datax)^T\textbf{A}^{-1}\g(\datax),
\end{equation}
where we have now explicitly stated the dependency of $\sigma^2_{\dset}(\datax)$ on $\datax$. The right side of equation \ref{Equation::post_pred_y_ready} follows straightforwardly using known results on multidimensional Gaussian integrals. Also, the right side of equation \ref{Equation::post_pred_variance_result} results conveniently via algebraic manipulation. Detailed results for equations \ref{Equation::post_pred_y_ready} and \ref{Equation::post_pred_variance_result} can be found from the appendix of this manuscript. By now simply discarding the factor $(2\pi)^{q/2}\lvert\textbf{A}+\beta\g\g^T\rvert^{-1/2}$ from the right side of equation \ref{Equation::post_pred_y_ready} and adding a multiplying factor $\{2\pi\sigma^2_{\dset}(\datax)\}^{-1/2}$, we can write $p(y|\datax, \dset)$ as: 
\begin{equation}
p(y|\datax, \dset)=\frac{1}{\sqrt{2\pi\sigma^2_{\dset}(\datax)}}\exp{\left(-\frac{\{y-\ymean\}^2}{2\sigma^2_{\dset}(\datax)}\right)}. 
\label{Equation::post_pred_y_equation_complete}
\end{equation}
One might have an issue with dropping out the non-constant factor $t(\datax)\overset{\Delta}{=}(2\pi)^{q/2}\lvert\textbf{A}+\beta\g(\datax)\g(\datax)^T\rvert^{-1/2}$ in equation \ref{Equation::post_pred_y_ready} but this is not a problem, since it simply scales the distribution function of $y\lvert \datax, \dset$ and the variance $\sigma^2_{\dset}(\datax)$ is invariant to this effect. Regarding the Algorithm \ref{Algorithm::Bayesian_sampling}, the factor $t(\datax)$ is also irrelevant and does not affect the functionality of BMVI sampling. Thus, we have now that the conditional posterior predictive distribution of $y$, given an input datum $\datax$ and data set $\dset$ is $y\lvert \datax, \dset \sim \mathcal{N}(\ymean, \sigma^2_{\dset}(\datax))$.

In a special case, if we use a linear function as the prediction model, i.e. $f(\datax; \w) = \datax^T\w + \theta_0$, then it is easy to show that the Hessian matrix $\textbf{A}$ of $S(\w)$ in equation \ref{Equation::error_function_S(w)} has the form:
\begin{equation}
\textbf{A} = \beta X^T X + \alpha \begin{bmatrix} 0 & \textbf{0}_{1\times m} \\
\textbf{0}_{m\times 1} & I_{m\times m} 
\end{bmatrix},
\end{equation}
where $\textbf{0}_{m\times 1}$ and $\textbf{0}_{1\times m}$ are $m$-dimensional zero vectors and matrix $X$ is defined as:
\begin{equation*}
X = \begin{pmatrix}
    1       & x_{11} & x_{12} & \dots & x_{1m} \\
    1       & x_{21} & x_{22} & \dots & x_{2m} \\
    \vdots  & \vdots & \vdots & \ddots & \vdots \\
    1       & x_{N1} & x_{N2} & \dots & x_{Nm}
\end{pmatrix},
\end{equation*}
where $i^{\text{th}}$ row contains the $i^{\text{th}}$ input vector $\datax_i$ (with term $1$ corresponding to constant parameter $\theta_0$). We see that $\textbf{A}$ is a positive semidefinite matrix, implying the convexity of $S(\w)$. This means the maximum posterior point $\wmp$ for a linear model is: 
\begin{equation*}
\wmp = \argmin_{\w\in \mathbb{R}^m, \theta_0\in\mathbb{R}} S(\w) = \left(X^TX + \frac{\alpha}{\beta}\begin{bmatrix} 0 & \textbf{0}_{1\times m} \\
\textbf{0}_{m\times 1} & I_{m\times m} 
\end{bmatrix}\right)^{-1}X^T\textbf{y},
\end{equation*}
where $\textbf{y}$ is a $N\times 1$ vector of output values. It follows that the variance of $p(y|\datax, \dset)$ for a linear prediction model is: 
\begin{equation}
\sigma^2_{\dset}(\datax)= \frac{1}{\beta} +\datax^T\textbf{A}^{-1}\datax.
\label{Equation::linear_model_predictive_variance}
\end{equation}

\pagebreak

Finally, we will present the derivations of the results in equations \ref{Equation::post_pred_y_ready} and \ref{Equation::post_pred_variance_result}. In the following equations, we denote $\textbf{C}=\textbf{A}+\beta\g\g^T$ and $D = \g^T \textbf{C}^{-1}\g$. We will also take advantage of the following known results: 
\begin{equation*}
\centering
\begin{gathered}
    \int_{-\infty}^{\infty} \exp{\left(-\frac{\lambda}{2}x^2\right)}\,dx = \left(\frac{2\pi}{\lambda}\right)^{1/2}\\ 
    \int_{\mathbb{R}^q}\exp{\left(-\frac{1}{2}\w^T \textbf{A}\w + \textbf{h}^T\w\right)\, d\w} = (2\pi)^{q/2}\lvert\textbf{A}\rvert^{-1/2}\exp{\left(\frac{1}{2}\textbf{h}^T\textbf{A}^{-1}\textbf{h}\right)}
    \end{gathered},
\end{equation*}
where $\textbf{A}$ is a real symmetric matrix, $\textbf{h}$  and $\w$ are $q$-dimensional vectors, and the integration is over whole $\w$-space $\mathbb{R}^q$.

\subsubsection*{Equation \ref{Equation::post_pred_y_ready}, closed form of $p(y|\datax, \dset)$:}
{\small
\begin{equation*}
\begin{aligned}
\label{Equation::pyxd_over_w_integral}
\pyxd \;&\propto\;  \infmint \exp{\left(-\frac{\beta}{2}\left\{y-\ymean-\g^T \w\right\}^2-\frac{1}{2}\w^T \textbf{A}\w\right)}\; d\w\\
&= \infmint \exp\left(-\frac{\beta}{2}\{y-\ymean\}^2+\beta \{y-\ymean\}\g^T\w -\frac{\beta}{2} \w^T\g\g^T\w-\frac{1}{2}\w^T\textbf{A}\w\right)\;d\w\\
&= \exp\left(-\frac{\beta}{2}\{y-\ymean\}^2\right)\infmint \exp\left(-\frac{1}{2}\w^T\textbf{C}\w+\beta \{y-\ymean\}\g^T\w\right)\;d\w\\
&= \exp\left(-\frac{\beta}{2}\{y-\ymean\}^2\right) \left[(2\pi)^{q/2}|\textbf{C}|^{-1/2} \exp\left(\frac{1}{2}\beta \{y-\ymean\}\g^T\textbf{C}^{-1}\beta \{y-\ymean\}\g\right)\right]\\
&= \exp\left(-\frac{\beta}{2}\{y-\ymean\}^2\right) \left[(2\pi)^{q/2}|\textbf{C}|^{-1/2} \exp\left(\frac{1}{2}\beta^2 \{y-\ymean\}^2 D\right)\right]\\
&=(2\pi)^{q/2}|\textbf{C}|^{-1/2} \exp\left(-\frac{\beta}{2}\{y-\ymean\}^2\right)  \exp\left((-D\beta)\left(-\frac{\beta}{2}\right) \{y-\ymean)^2 \right)\\
&= (2\pi)^{q/2}|\textbf{C}|^{-1/2}    \exp\left((1-D\beta)\left(-\frac{\beta}{2}\right) \{y-\ymean\}^2 \right)\\
&= (2\pi)^{q/2}|\textbf{A}+\beta\g\g^T|^{-1/2} \exp\left(-\frac{\beta-\beta^2\g^T (\textbf{A}+\beta\g\g^T)^{-1}\g}{2} \{y-\ymean\}^2 \right)\\
&= (2\pi)^{q/2}|\textbf{A}+\beta\g\g^T|^{-1/2} \exp\left(-\frac{\{y-\ymean\}^2}{2\sigma_{\dset}^2(\datax)} \right), 
\end{aligned}
\end{equation*}
}
where $\sigma_{\dset}^2(\datax) = \left(\beta-\beta^2\g^T (\textbf{A}+\beta\g\g^T)^{-1}\g\right)^{-1}\;\;\blacksquare\;$ Note that integrating $\pyxd$ with respect to $y$ gives: 
\begin{equation*}
\begin{aligned}
\label{Equation::pyxd_over_t_integral}
\infint\, \pyxd\; dy\; &\propto \infint (2\pi)^{q/2}|\textbf{A}+\beta\g\g^T|^{-1/2} \exp\left(-\frac{\{y-\ymean\}^2}{2\sigma_{\dset}^2(\datax)} \right)\;dy\\
&= (2\pi)^{q/2}|\textbf{A}+\beta\g\g^T|^{-1/2} \infint \exp\left(-\frac{\{y-\ymean\}^2}{2\sigma_{\dset}^2(\datax)} \right)\;dy\\
&= (2\pi)^{q/2}|\textbf{A}+\beta\g\g^T|^{-1/2}\sqrt{2\pi\sigma_{\dset}^2(\datax)},
\end{aligned}
\end{equation*}
which contains the reciprocal of $\{2\pi\sigma_{\dset}^2(\datax)\}^{-1/2}$ we added in equation \ref{Equation::post_pred_y_equation_complete}. 

\subsubsection*{Equation \ref{Equation::post_pred_variance_result}, closed form of $\yvar$:}
{\small
\begin{equation*}
\begin{aligned}
\sigma^2_\mathcal{D}(\datax) &= \frac{1}{\beta-\beta^2\g^T(\textbf{A}+\beta\g\g^T)^{-1}\g}\\
&= \frac{1}{\beta-\beta^2\g^T(\textbf{A}+\beta\g\g^T)^{-1}\g}\times\frac{\g^T(\textbf{I}+\beta\textbf{A}^{-1}\g\g^T)\g}{\g^T(\textbf{I}+\beta\textbf{A}^{-1}\g\g^T)\g}\\
&= \frac{\g^T\g+\g^T\beta\textbf{A}^{-1}\g\g^T\g}{\beta\g^T\g+\beta^2\g^T\textbf{A}^{-1}\g\g^T\g-\beta^2\g^T(\textbf{A}+\beta\g\g^T)^{-1}\g\g^T\g-\beta^2\g^T(\textbf{A}+\beta\g\g^T)^{-1}\g\g^T(\beta\textbf{A}^{-1}\g\g^T)\g}\\
&=\frac{1+\g^T\beta\textbf{A}^{-1}\g}{\beta+\beta^2\g^T\textbf{A}^{-1}\g-\beta^2\g^T(\textbf{A}+\beta\g\g^T)^{-1}\g-\beta^2\g^T(\textbf{A}+\beta\g\g^T)^{-1}\g\g^T\beta\textbf{A}^{-1}\g}\\
&=\frac{\frac1\beta+\g^T\textbf{A}^{-1}\g}{1+\beta\g^T\textbf{A}^{-1}\g-\beta\g^T(\textbf{A}+\beta\g\g^T)^{-1}\g-\beta\g^T(\textbf{A}+\beta\g\g^T)^{-1}\g\g^T\beta\textbf{A}^{-1}\g}.
\end{aligned}
\end{equation*}
}
\noindent Now in order for the result in equation \ref{Equation::post_pred_variance_result} to hold, the denominator excluding term $1$ in the above fraction should equate to $0$: 
\begin{equation*}
\begin{aligned}
\beta\g^T\textbf{A}^{-1}\g-\beta\g^T(\textbf{A}+\beta\g\g^T)^{-1}\g-\beta\g^T(\textbf{A}+\beta\g\g^T)^{-1}\g\g^T\beta\textbf{A}^{-1}\g&=0\\
\g^T\textbf{A}^{-1}\g-\g^T(\textbf{A}+\beta\g\g^T)^{-1}\g-\g^T(\textbf{A}+\beta\g\g^T)^{-1}\g\g^T\beta\textbf{A}^{-1}\g&=0
\end{aligned}    
\end{equation*}
\begin{equation*}
\begin{aligned}
\g^T\textbf{A}^{-1}\g-\g^T(\textbf{A}+\beta\g\g^T)^{-1}\g &=\g^T(\textbf{A}+\beta\g\g^T)^{-1}\g\g^T\beta\textbf{A}^{-1}\g\\
\g^T\textbf{A}^{-1}\g-\g^T\textbf{C}^{-1}\g &= \beta\g^T\textbf{C}^{-1}\g\g^T\textbf{A}^{-1}\g\\
\g^T\textbf{A}^{-1}\g\g^T(\g\g^T)^{-1}\textbf{C}-\g^T\textbf{C}^{-1}\g\g^T(\g\g^T)^{-1}\textbf{C} &= \beta\g^T\textbf{C}^{-1}\g\g^T\textbf{A}^{-1}\g\g^T(\g\g^T)^{-1}\textbf{C}\\
\g^T\textbf{A}^{-1}\textbf{C}-\g^T &= \beta\g^T\textbf{C}^{-1}\g\g^T\textbf{A}^{-1}\textbf{C}\\
\textbf{C}(\g\g^T)^{-1}\g\g^T\textbf{A}^{-1}\textbf{C}-\textbf{C}(\g\g^T)^{-1}\g\g^T &= \beta\textbf{C}(\g\g^T)^{-1}\g\g^T\textbf{C}^{-1}\g\g^T\textbf{A}^{-1}\textbf{C}\\
\textbf{C}\textbf{A}^{-1}\textbf{C}-\textbf{C} &= \beta\g\g^T\textbf{A}^{-1}\textbf{C}\\
(\textbf{A}+\beta\g\g^T)\textbf{A}^{-1}(\textbf{A}+\beta\g\g^T)-(\textbf{A}+\beta\g\g^T) &= \beta\g\g^T\textbf{A}^{-1}(\textbf{A}+\beta\g\g^T)\\
(\textbf{A}+\beta\g\g^T)(\textbf{I}+\beta\textbf{A}^{-1}\g\g^T)-(\textbf{A}+\beta\g\g^T) &= \beta(\g\g^T+\beta\g\g^T\textbf{A}^{-1}\g\g^T)\\
(\textbf{A}+\beta\g\g^T)(\beta\textbf{A}^{-1}\g\g^T) &= \beta(\g\g^T+\beta\g\g^T\textbf{A}^{-1}\g\g^T)\\
\beta(\g\g^T+\beta\g\g^T\textbf{A}^{-1}\g\g^T)&=\beta(\g\g^T+\beta\g\g^T\textbf{A}^{-1}\g\g^T).
\end{aligned}    
\end{equation*}

In other words, 

$$\sigma^2_{\dset}(\datax) = \frac{1}{\beta-\beta^2\g^T(\textbf{A}+\beta\g\g^T)^{-1}\g}=\frac1\beta+\g^T\textbf{A}^{-1}\g \;\;\;\blacksquare$$

\end{document}